\documentclass[pdflatex,sn-nature]{sn-jnl}% Style for submissions to Nature Portfolio journals
\usepackage{graphicx}%
\usepackage{multirow}%
\usepackage{amsmath,amssymb,amsfonts}%
\usepackage{amsthm}%
\usepackage{mathrsfs}%
\usepackage[title]{appendix}%
\usepackage{xcolor}%
\usepackage{textcomp}%
\usepackage{manyfoot}%
\usepackage{booktabs}%
\usepackage{algorithm}%
\usepackage{algorithmicx}%
\usepackage{algpseudocode}%
\usepackage{listings}%
\theoremstyle{thmstyleone}%
\theoremstyle{thmstyletwo}%

\theoremstyle{thmstylethree}%

\usepackage{hyperref}
\usepackage{url}
\usepackage{graphicx}
\usepackage{multirow}
\usepackage{booktabs}
\usepackage{array}
\usepackage{caption}
\usepackage[inkscapelatex=false]{svg}
\usepackage{comment}
\usepackage{tabularx}
\usepackage{marginnote}

\newcommand{\JambaDNA}{HybriDNA}

\begin{document}

% \title{HybriDNA: A Hybrid MAMBA2-Transformer Model for DNA Understanding and Generation}
\title{HybriDNA: A Hybrid Transformer-Mamba2 Long-Range DNA Language Model}

% Authors must not appear in the submitted version. They should be hidden
% as long as the \iclrfinalcopy macro remains commented out below.
% Non-anonymous submissions will be rejected without review.

\author[1,2]{\fnm{Mingqian} \sur{Ma}}\email{mingqianma@sjtu.edu.cn}
\equalcont{These authors contributed equally to this work.}

\author[1]{\fnm{Guoqing } \sur{Liu}}\email{guoqingliu@microsoft.com}
\equalcont{These authors contributed equally to this work.}

\author[1]{\fnm{Chuan} \sur{Cao}}\email{chuancao.926@gmail.com}
\equalcont{These authors contributed equally to this work.}

\author[1]{\fnm{Pan} \sur{Deng}}\email{pan.deng@microsoft.com}
\equalcont{These authors contributed equally to this work.}

\author[3]{\fnm{Tri} \sur{Dao}}\email{tri@tridao.me}

\author[4]{\fnm{Albert} \sur{Gu}}\email{agu@andrew.cmu.edu}

\author[1]{\fnm{Peiran} \sur{Jin}}\email{peiranjin@microsoft.com}

\author[5]{\fnm{Zhao} \sur{Yang}}\email{yangyz1230@gmail.com}

\author[1]{\fnm{Yingce} \sur{Xia}}\email{yingce.xia@microsoft.com}

\author[1]{\fnm{Renqian} \sur{Luo}}\email{renqianluo@microsoft.com}

\author[1]{\fnm{Pipi} \sur{Hu}}\email{pisquare@microsoft.com}

\author[1]{\fnm{Zun} \sur{Wang}}\email{zunwang@microsoft.com}

\author[1]{\fnm{Yuan-Jyue} \sur{Chen}}\email{yuanjc@microsoft.com}

\author[1]{\fnm{Haiguang} \sur{Liu}}\email{haiguang.liu@microsoft.com}

\author*[1]{\fnm{Tao} \sur{Qin}}\email{taoqin@microsoft.com}

\affil[1]{Microsoft Research AI for Science}

\affil[2]{UM-SJTU Joint Institute, Shanghai Jiao Tong University}

\affil[3]{\orgdiv{Department of Computer Science}, \orgname{Princeton University}}

\affil[4]{\orgdiv{Machine Learning Department}, \orgname{Carnegie Mellon University}}

\affil[5]{\orgdiv{Gaoling School of Artificial Intelligence}, \orgname{Renmin University of China}}
\maketitle
% The \author macro works with any number of authors. There are two commands
% used to separate the names and addresses of multiple authors: \And and \AND.
%
% Using \And between authors leaves it to \LaTeX{} to determine where to break
% the lines. Using \AND forces a linebreak at that point. So, if \LaTeX{}
% puts 3 of 4 authors names on the first line, and the last on the second
% line, try using \AND instead of \And before the third author name.
% \newcommand{\fix}{\marginpar{FIX}}
% \newcommand{\new}{\marginpar{NEW}}
\begin{abstract}
Advances in natural language processing and large language models have sparked growing interest in modeling DNA, often referred to as the ``language of life". 
However, DNA modeling poses unique challenges. 
First, it requires the ability to process ultra-long DNA sequences while preserving single-nucleotide resolution, 
as individual nucleotides play a critical role in DNA function.
% Second, while generative tasks are increasingly valued for their therapeutic and industrial potential,
% understanding DNA remains equally vital, 
% as it provides insights into biological mechanisms and diseases.
Second, success in this domain requires excelling at both generative and understanding tasks: generative tasks hold potential for therapeutic and industrial applications, while understanding tasks provide crucial insights into biological mechanisms and diseases.
To address these challenges, we propose \textbf{\JambaDNA}, a decoder-only DNA language model that incorporates a hybrid Transformer-Mamba2 architecture, seamlessly integrating the strengths of attention mechanisms with selective state-space models.
% effectively combining the strengths of attention mechanisms with selective state-space models.
% This hybrid design enables \JambaDNA{} to process DNA sequences up to 131kb in length at single-nucleotide resolution with exceptional efficiency. 
This hybrid design enables \JambaDNA{} to efficiently process DNA sequences up to 131kb in length with single-nucleotide resolution.
% \JambaDNA{} achieves state-of-the-art performance across 35 DNA understanding datasets curated by the BEND, GUE, and LRB benchmarks, and it excels at generating synthetic cis-regulatory elements (CREs) with desired properties.
\JambaDNA{} achieves state-of-the-art performance across 33 DNA understanding datasets curated from the BEND, GUE, and LRB benchmarks, and demonstrates exceptional capability in generating synthetic cis-regulatory elements (CREs) with desired properties. 
Furthermore, we show that \JambaDNA{} adheres to expected scaling laws, with performance improving consistently as the model scales from 300M to 3B and 7B parameters.
% Furthermore, we demonstrate that \JambaDNA{} adheres to Scaling Laws, with performance improving as the model scales from 300M to 3B and 7B parameters. 
% These findings underscore \JambaDNA’s versatility and its potential to drive advancements in DNA research and applications.
These findings underscore \JambaDNA's versatility and its potential to advance DNA research and applications, paving the way for innovations in understanding and engineering the ``language of life".

\end{abstract}
\section{Introduction}
Deoxyribonucleic acid (DNA) serves as the genetic code of life, encoding the instructions that govern gene expression, cellular processes, and biological functions.
A deep understanding of the ``language'' of DNA is crucial for unraveling the molecular mechanisms that underlie biological functions and for leveraging these insights to advance medicine and biotechnology. 
The advent of high-throughput sequencing technologies has generated an immense volume of genomic data, creating an unprecedented opportunity for machine learning models to uncover complex patterns and relationships within DNA sequences. 
Foundation models, pretrained on large-scale unlabeled datasets, 
have already demonstrated remarkable capabilities in natural languages~\citep{devlin2018bert, bommasani2021opportunities,achiam2023gpt} and protein languages~\citep{brandes2022proteinbert, lin2023evolutionary, he2024sfm}.

Recently, foundation models have begun to drive a paradigm shift in genomics,  showcasing their ability to learn rich representations of DNA sequences that can be fine-tuned for a diverse array of downstream tasks.
Currently, DNA foundation models primarily adopt two main architectural approaches. 
The first approach, inspired by BERT~\citep{devlin2018bert}, employs encoder-only Transformer architectures. Models such as DNABERT2~\citep{zhou2023dnabert2} and Nucleotide Transformer (NT)~\citep{dalla2023nucleotide} excel at capturing contextual information within DNA sequences, producing high-quality embeddings suitable for tasks such as classification and regression. 
However, their bidirectional nature constrains their ability to design novel DNA sequences.
The second approach leverages decoder-only architectures, such as Hyena~\citep{poli2023hyena} and the Transformer architecture in GPT~\citep{radford2018improving}, which are autoregressive and well-suited for generative tasks. Models like HyenaDNA~\citep{poli2023hyenadna} and Evo~\citep{meier2023evo} have shown promising results in generating DNA sequences. 
Nevertheless, they often fall behind encoder-only models in understanding tasks requiring a deep understanding of sequence context.

This dichotomy highlights two critical challenges in DNA modeling:
(1) How to develop a DNA foundation model that integrates robust contextual understanding with advanced design capabilities?
Such a model would not only enhance the analysis of existing genomic data but also enable the design of novel, functional DNA sequences.
(2) How to efficiently address the intricate complexity of DNA sequences, which involves long-range interactions critical to fundamental biological processes?
Recent advances in Selective State Space Models (SSMs), such as Mamba~\citep{gu2023mamba, dao2024transformers}, have shown remarkable potential for addressing information-dense tasks, including language modeling~\citep{waleffe2024empirical,lieber2024jambahybridtransformermambalanguage}. 
These models efficiently handle long-range dependencies with subquadratic complexity, 
offering a promising approach to the challenges posed by DNA sequence modeling. However, SSMs alone struggle to capture fine-grained, single-nucleotide-level interactions vital for understanding DNA function.

In this work, we introduce \JambaDNA, a novel class of decoder-only DNA language models that leverage a hybrid Transformer-Mamba2 architecture. 
This hybrid design combines the complementary strengths of its components:
Mamba2 blocks excel at efficiently processing long sequences and capturing long-range dependencies, whereas Transformer blocks enhance the model's ability to focus on fine-grained, token-level details within the context of the entire sequence.
Pretrained on large-scale, multi-species genomes at single-nucleotide resolution with a next-token prediction objective, \JambaDNA{} demonstrates foundational capabilities in both understanding and designing genomic sequences.
By incorporating an \textit{echo embedding} discriminative fine-tuning approach,
\JambaDNA{} achieves state-of-the-art performance across 35 biologically significant DNA understanding datasets, such as transcription factor binding prediction and promoter detection~\citep{zhou2023dnabert2}.
Additionally, through generative fine-tuning, 
\JambaDNA{} exhibits exceptional proficiency in designing synthetic cis-regulatory elements (CREs) with desirable functional properties, such as yeast promoters and cell type-specific human enhancers~\citep{lal2024reglm}.
Finally, we show that scaling up \JambaDNA{} is beneficial: increasing model size from 300 million to 3 billion and 7 billion parameters improves performance, adhering to scaling laws observed in language models such as GPT~\citep{radford2018improving,schulman2022chatgpt}.
Extending the context length (e.g., from 8 kilobases to 131 kilobases at single-nucleotide resolution) further enhances \JambaDNA’s performance on specific tasks.
Together, these advancements position \JambaDNA{} as a powerful tool for advancing both the understanding and engineering of genomic sequences.

Our contributions are summarized as follows:
\begin{itemize}
    \item We propose \JambaDNA, 
     a class of decoder-only DNA language models that integrates a hybrid Transformer-Mamba2 architecture.
    Leveraging this architecture, we develop a comprehensive training pipeline that includes pretraining, echo embedding-based discriminative fine-tuning, and generative fine-tuning.
    \item  \JambaDNA{} achieves state-of-the-art performance across 33 diverse and biologically meaningful DNA understanding datasets, spanning human and multi-species genomes. \JambaDNA{} outperforms many existing encoder-only models.
    \item \JambaDNA{} demonstrates its generative capabilities, showcasing its proficiency in designing desirable synthetic cis-regulatory elements.
    \item \JambaDNA{} demonstrates the ability to process long-context DNA sequences on tasks involving the analysis of extended DNA sequences (e.g., 131 kilobases).
    \item Scaling laws for \JambaDNA{} are observed: increasing the model size from 300M to 3B and 7B parameters improves performance.
\end{itemize}

\section{Preliminaries}
\subsection{Attention Mechanism in Transformers}
\label{sec:attention}
Powering many foundation models is the attention mechanism~\citep{bahdanau2014neural,vaswani2017attention} in Transformers.
Attention is a type of operator that assigns scores to every pair of tokens in a sequence, enabling each element to ``attend'' to the others.
% The most widely used variant of attention so far 
The most widely adopted variant of attention to date is Scaled Dot-Product Attention, which is defined as:
\begin{equation}
    y = \mathrm{softmax}(\frac{QK^{T}}{\sqrt{d_{k}}})\cdot V.
    \label{eqn:attention}
\end{equation}

Let $x \in \mathbb{R}^{L \times d}$ represents an input sequence with sequence length $L$ and embedding size $d$, the learnable parameters $W_{K} \in \mathbb{R}^{d \times d_k}, W_{Q} \in \mathbb{R}^{d \times d_k}$, and $W_{V} \in \mathbb{R}^{d \times d}$ are used to compute the key, query, and value matrices: $K = xW_{K}$, $Q = xW_{Q}$, and $V = xW_{V}$. 
The attention layer, therefore, transforms an input $x$ of shape $\mathbb{R}^{L \times d}$ into an output $y$ of the same shape, $\mathbb{R}^{L \times d}$.

Attention computes all pairwise comparisons for every token in a sequence, resulting in a computational complexity that scales as $O(L^2)$ with sequence length $L$. While this enables capturing global context at high resolution, it also restricts the context length on modern GPU architectures.

\subsection{Selective State Space Models}

% State Space Models (SSMs) and Mamba are inspired by the continuous mapping of a one-dimensional function \( f: x(t) \mapsto y(t) \) through a hidden state representation \( h(t) \in \mathbb{R}^N \):

Structured state space sequence models (S4) \citep{gu2022efficiently, gu2021combining} are a recent class of sequence models in deep learning that are broadly related
to RNNs, CNNs, and classical state space models. 
They are inspired by a specific continuous system, which maps a 1-dimensional sequence $x \in \mathbb{R}^{L} \mapsto y \in \mathbb{R}^{L}$ via a hidden state \( h \in \mathbb{R}^{(L,N)} \), where $L$ represents the sequence length, and $N$ represents the SSM state size. 

Specifically, the continuous system is defined by three matrices $(\mathbf{A}, \mathbf{B}, \mathbf{C})$,
$\mathbf{A} \in \mathbb{R}^{N \times N}$, $\mathbf{B} \in \mathbb{R}^{N \times 1}$, and $\mathbf{C} \in \mathbb{R}^{1 \times N}$.
 They define a sequence-to-sequence transformation in two steps, as detailed in Eqn.~\ref{eqn:ssm} (first column).
% \begin{equation}
%     h_t = A h_{t-1} + B x_t, \quad y_t = C h_t
% \end{equation}
\begin{equation}  
\begin{aligned}  
h'(t) &= \mathbf{A} h(t) + \mathbf{B} x(t), &\quad h_t &= \mathbf{\bar{A}} h_{t-1} + \mathbf{\bar{B}} x_t, &\quad  \mathbf{\bar{K}} &= (\mathbf{C}\mathbf{\bar{B}},\mathbf{C}\mathbf{\bar{A}}\mathbf{\bar{B}},...,\mathbf{C}\mathbf{\bar{A}}^{k}\mathbf{\bar{B}}),  \\  
y(t) &= \mathbf{C} h(t). &\quad  y_t &= \mathbf{C}h_t. &\quad  y&=x*\mathbf{\bar{K}}.  
\end{aligned}
\label{eqn:ssm}
\end{equation}  

\textbf{Discretization}~S4 models are discrete versions of the continuous system (as shown in the second column of Eqn.~\ref{eqn:ssm}), which incorporates a timescale parameter $\Delta \in \mathbb{R}$ to transform the continuous parameters $\mathbf{A}$, and $\mathbf{B}$
into their discrete counterparts, $\mathbf{\bar{A}}$ and $\mathbf{\bar{B}}$. 
% A commonly used transformation for this process is the zero-order hold (ZOH), defined as follows, where ``exp'' denotes the exponential:
The zero-order hold (ZOH) is a commonly used method for this transformation, defined as follows, where "exp" represents the exponential function.
% By discretizing in the time domain using a rule like the \textit{zero-order hold (ZOH)}, the matrices \( A \) and \( B \) are transformed into their discrete counterparts:
\begin{equation}
    \mathbf{\bar{A}} = \exp(\Delta \mathbf{A}), \quad \mathbf{\bar{B}} = (\Delta \mathbf{A})^{-1}(\exp(\Delta \mathbf{A}) - \mathbf{I}) \cdot \Delta \mathbf{B}.
    \label{eqn:discrete_ssm}
\end{equation}

% which serves as an important component in models like H3 \citep{fu2023hungryhungryhipposlanguage}.

\textbf{Computation}~After the parameters have been transformed into $(\mathbf{\bar{A}}, \mathbf{\bar{B}})$, the model can be computed in two ways, either as a \underline{linear recurrence} (corresponding to
the second column of Eqn.~\ref{eqn:ssm}) or a \underline{global convolution} (corresponding to
the third column of Eqn.~\ref{eqn:ssm}). 
Commonly, the model uses the convolutional mode for efficient parallelizable training (where the whole input sequence is seen ahead of time), and switched into recurrent mode for efficient autoregressive inference (where the inputs are
seen one timestep at a time).

\textbf{Structured matrix $\mathbf{A}$} S4 models are named as such because efficiently computing them requires imposing a specific structure on the $\mathbf{A}$ matrix. The most popular structure is diagonal~\citep{gupta2022diagonal,gu2022parameterization,smith2023simplified}.
In this scenario, the $\mathbf{A}, \mathbf{B}, \mathbf{C}$ matrices can each be represented by $N$ numbers. To process an input sequence $x$ of $D$ channels, the SSM is applied independently to each channel.

\textbf{Linear Time Invariance (LTI)}~A key characteristic of Eqn.~\ref{eqn:ssm} is its linear time invariance, meaning the model's dynamics remain constant over time.
% that the model’s dynamics remain constant over time, a property known as Linear Time Invariance.
In other words, the matrices $(\Delta, \mathbf{A}, \mathbf{B}, \mathbf{C})$, and consequently $(\mathbf{\bar{A}}, \mathbf{\bar{B}})$, are fixed for all time-steps. 
The LTI property is closely associated with recurrence and convolutions. 
Before Mamba, all S4 models adhered to LTI, often computed as convolutions during training.
% Before Mamba, all S4 models adhered to LTI (e.g. computed as convolutions during training).
% because of fundamental efficiency constraints.

% This above formulation is central to S4. 
H3~\citep{fu2022hungry} generalizes the recurrence to use S4; it can be viewed as an architecture with an SSM sandwiched by two gated connections. H3 also inserts a standard local convolution, which they frame as a
shift-SSM, before the inner SSM layer.

\textbf{Mamba1}~\citep{gu2023mamba} introduces the concept of Selective SSMs~(S6), enhancing the traditional S4 framework through input-dependent gating mechanisms. The matrices $\mathbf{\bar A}, \mathbf{\bar B}$, and $\Delta$ are dynamically gated by the input $x_t$, enabling them to adjust their behavior based on the current input.
Mamba1 simplifies the block design by combining the H3 block~\cite{fu2023hungryhungryhipposlanguage} with gated MLPs.
% This gating mechanism allows the model to modulate the influence of the previous hidden state \( h_{t-1} \)  in response to \( x_t \).
% thereby improving its performance on tasks requiring retrieval and selective copying.
Additionally, Mamba1 proposes selective scan, a hardware-aware algorithm that computes the model recurrently using a scan operation, enhancing computational efficiency and scalability.

\textbf{Mamba2}~\citep{dao2024transformers} enhances Mamba1 by introducing two key enhancements: 

1. At the SSM layer, the new Structured State Space Duality (SSD) layer imposes a stricter constraint on the diagonal matrix $\mathbf{\bar{A}}$. The diagonal matrix is now simplified to a scalar times an identity matrix, 
allowing it to be represented using a single identical value across the diagonal. Consequently, $\mathbf{\bar{A}}$ can be represented with a shape corresponding only to the sequence length.
% which can be represented using only a single identical value across the diagonal. 
% In this case, $\mathbf{A}$ can be represented with shape just sequence length.
% This simplification leads to the following formulation:

% \begin{equation} \label{eq:ssm-transformation}
% y^\mathtt{(L,P)} = \mathsf{SSM}(\mathbf{A}^\mathtt{(L)}, \mathbf{B}^\mathtt{(L,N)}, \mathbf{C}^\mathtt{(L,N)})(x^\mathtt{(L,P)}),
% \end{equation}

% where $x \in \mathbb{R}^{L \times P}$ with sequence length $L$ and separate channel number $P$, we can use the same dynamics (i.e. the same SSM$(\mathbf{A},\mathbf{B},\mathbf{C})$ independently for each channel.
% In Mamba2, channel number $P$ is usually set to $64$ or $128$. 
% This can be interpreted as a single head of the SSM model.
% To scale to larger model widths, we keep this fixed and increase the number of independent heads.

2. The Mamba2 block produces the SSM parameters $(\mathbf{\bar{A}}, \mathbf{\bar{B}}, \mathbf{C})$  in parallel with the input $x$, as opposed to sequentially in the Mamba1 block. 
% This modification enables greater parallelism and scalability improvements, making tensor parallelism feasible for scaling the model to larger dimensions and longer contexts.
% Compared to Mamba, Mamba2 allows much larger state dimensions (from $N=16$ in Mamba1 to $N=64$ to $N=256$ or even higher) while simultaneously being much faster during training.
This change facilitates greater parallelism and scalability, enabling tensor parallelism for scaling the model to larger dimensions and longer contexts. Compared to Mamba1, Mamba2 supports much larger state dimensions (increasing from $N=16$ in Mamba1 to $N=64$, $N=256$, or even higher) while also significantly improving training speed.

\section{\JambaDNA{} Foundation Model}

In this section, we introduce the \JambaDNA{} model for long-range genomic sequence modeling. 
% We first provide a detailed description of the model architecture, followed by an explanation of the pretraining stage of \JambaDNA{}. Finally, we explore the fine-tuning stages utilized for a range of downstream applications. The pipeline of our model is illustrated in Fig. \ref{fig:pipeline}. 
We begin with a detailed description of the model architecture, followed by an explanation of the pretraining stage of HybriDNA. Finally, we discuss the fine-tuning stages used for various downstream applications. The model's pipeline is illustrated in Fig.~\ref{fig:pipeline}.

\subsection{Model Architecture}

% The \JambaDNA{} model is a decoder-only, sequence-to-sequence architecture designed to handle long-range DNA sequences with efficiency and accuracy. It combines the unique strengths of Mamba2 selective state-space models and Transformer attention mechanisms, interleaved within a hybrid framework inspired by Jamba~\citep{lieber2024jambahybridtransformermambalanguage}. 
% As illustrated in Fig.~\ref{fig:pipeline}, \JambaDNA{} architecture comprised of a seris of \JambaDNA{} blocks, which interleaves Mamba2 blocks and Transformer blocks in a 7:1 ratio, which is a optimial configuration has been demonstrated to effectively balance the strengths of both block types, resulting in optimal performance in general domains~\citep{waleffe2024empirical}.

% The \JambaDNA{} model is a decoder-only, sequence-to-sequence architecture specifically designed to efficiently and accurately process long-range DNA sequences. 
% It leverages the unique strengths of Mamba2 selective state-space models and Transformer attention mechanisms within a hybrid framework inspired by Jamba~\citep{lieber2024jambahybridtransformermambalanguage}. 
% As shown in Fig.\ref{fig:detail_architect}, the \JambaDNA{} architecture consists of a series of \JambaDNA{} blocks, which alternate HybriDNA Mamba2 blocks and HybriDNA Transformer blocks in a 7:1 ratio. This configuration has been empirically shown to effectively balance the advantages of both block types, achieving optimal performance in the NLP domain~\citep{waleffe2024empirical}.

\begin{figure}[t]
    \centering
    \centerline{\includegraphics[width=1.0\columnwidth]{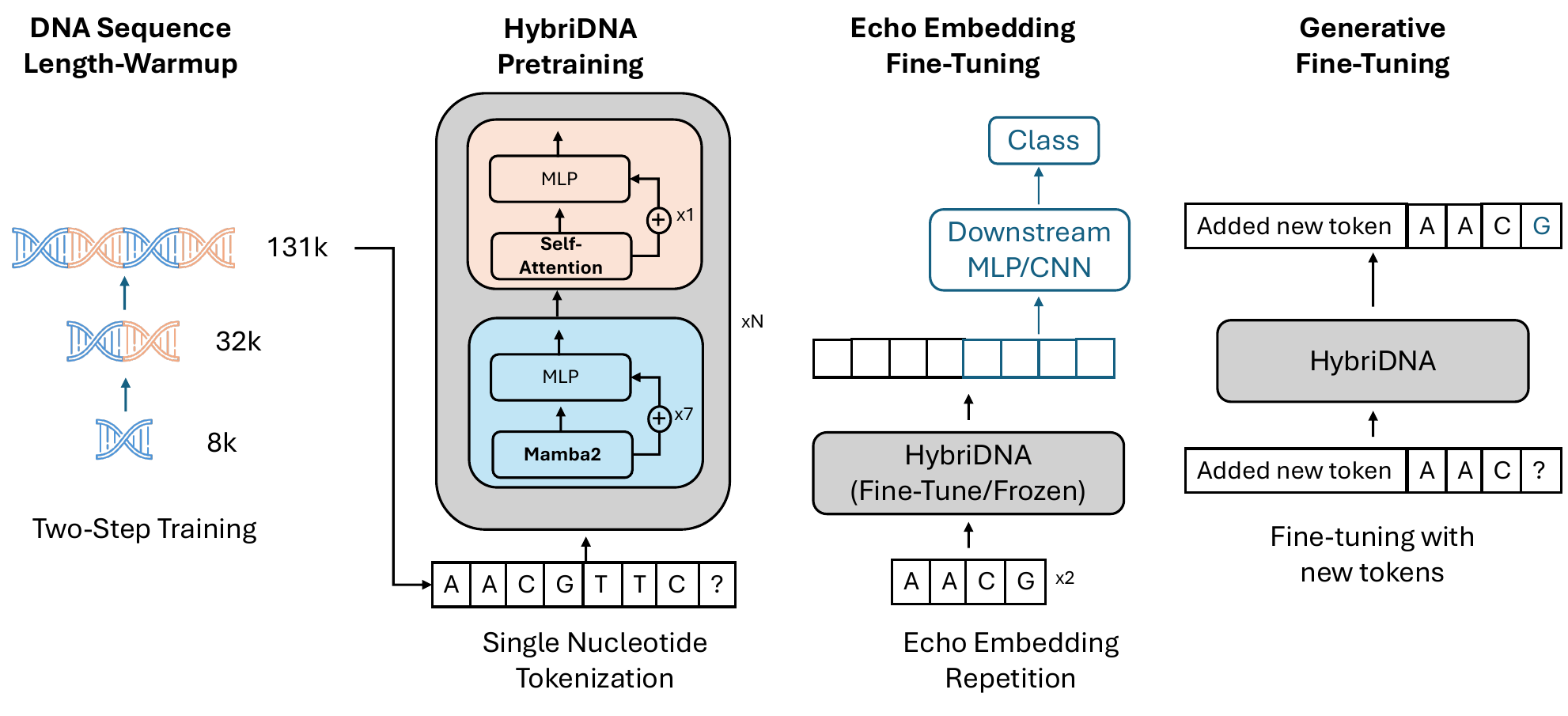}}
    \hspace{1cm}
    \caption{Overview of \JambaDNA{}: A Language Model for DNA Sequences. \JambaDNA{} builds upon an efficient hybrid Transformer and Mamba2 architecture. It is initially pretrained on large-scale, multi-species genomic data at single-nucleotide resolution using a next-token prediction objective. Subsequently, \JambaDNA{} employs an echo embedding fine-tuning approach for DNA understanding tasks and a generative fine-tuning approach for DNA generation tasks.}
    \label{fig:pipeline}
\end{figure}

The \JambaDNA{} model uses a decoder-only, sequence-to-sequence architecture purpose-built for efficiently and accurately processing long-range DNA sequences. 
It combines the strengths of Mamba2 selective state-space models and Transformer attention mechanisms within a hybrid framework inspired by recent hybrid architectures ~\citep{lieber2024jambahybridtransformermambalanguage,glorioso2024zambacompact7bssm,ren2024sambasimplehybridstate}. As shown in Fig.~\ref{fig:pipeline} (second column), the architecture consists of a series of \JambaDNA{} blocks, where each block alternates between \JambaDNA{} Mamba2 blocks and \JambaDNA{} Transformer blocks in a 7:1 ratio.
This configuration has been empirically proven to effectively balance the advantages of both block types, achieving optimal performance in the NLP domain~\citep{waleffe2024empirical}.

A key component of the \JambaDNA{} Mamba2 block is the \textbf{State-Space Duality (SSD)} layer \citep{dao2024transformers}. 
It processes input sequence $x$ using the recurrence:
\begin{equation}
    h_t = A_t h_{t-1} + B_t x_t, \quad y_t = C_t^\top h_t,
\end{equation}
where \(h_t \in \mathbb{R}^N\) denotes the hidden state, \(A_t \in \mathbb{R}^{N \times N}\) represents the state transitions, \(x_t \in \mathbb{R}\) is the input, \(B_t \in \mathbb{R}^{N \times 1}\) projects the input, and \(C_t \in \mathbb{R}^{N \times 1}\) maps the hidden state to the output \(y_t \in \mathbb{R}\).

The SSD layer simplifies the matrix \(A_t\) to \(A_t = a_t I\), where \(a_t \in \mathbb{R}\) is a scalar and \(I\) is the identity matrix, to further improve efficiency. This simplification reduces the recurrence to:
\begin{equation}
h_t = a_t h_{t-1} + B_t x_t.
\end{equation}

For multi-dimensional (channel) inputs $x \in \mathbb{R}^{L \times d}$, the SSD layer is extended into a multi-head design, where each head independently processes a distinct subset of the input dimensions, similar to the multi-head attention mechanism in Transformers. 
This architecture allows the SSD layer to capture complex interactions across multiple input channels in parallel, greatly enhancing its representational capacity. 
Typically, the head dimension is set to 64 or 128, consistent with configurations used in standard transformers.

\begin{figure}[t]
\centering
\includegraphics[width=0.8\textwidth]{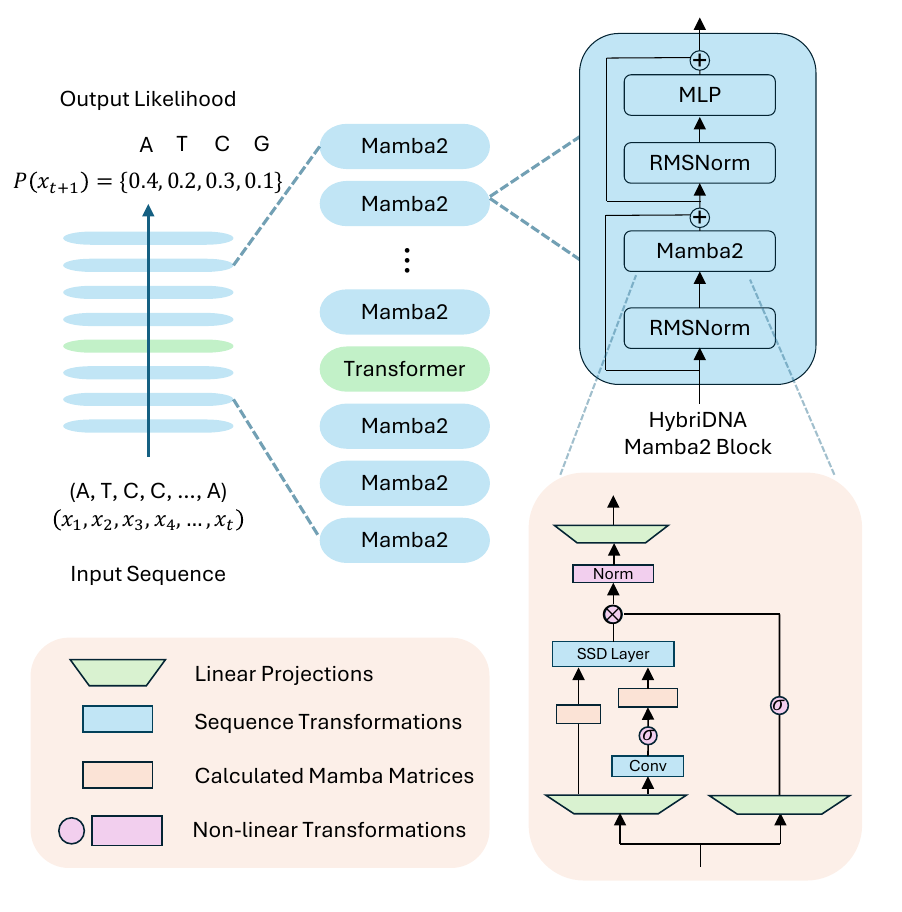}
\caption{Model Architecture of \JambaDNA{}}
\label{fig:detail_architect}
\end{figure}

% A major component of the Mamba2 block in \JambaDNA{} is the \textbf{State-Space Duality (SSD)} layer \citep{dao2024transformers}. 
% It processes input sequence $x$ efficiently using the recurrence:
% \begin{equation}
%     h_t = A_t h_{t-1} + B_t x_t, \quad y_t = C_t^\top h_t,
% \end{equation}
% where \(h_t \in \mathbb{R}^N\) is the hidden state, \(x_t \in \mathbb{R}\) is the input, \(A_t \in \mathbb{R}^{N \times N}\) represents state transitions, \(B_t \in \mathbb{R}^{N \times 1}\) projects the input, and \(C_t \in \mathbb{R}^{N \times 1}\) maps the hidden state to the output \(y_t \in \mathbb{R}\).

% To improve efficiency, the SSD layer simplifies the matrix \(A_t\) to \(A_t = a_t I\), where \(a_t \in \mathbb{R}\) is a scalar and \(I\) is the identity matrix. This simplification reduces the recurrence to:
% \begin{equation}
% h_t = a_t h_{t-1} + b_t x_t, \quad b_t = B_t.
% \end{equation}

Computationally, the SSD layer can be reformulated as a matrix operation:
\[
y = Mx, \quad M_{ij} =
\begin{cases}
C_i^\top A_{j:{i+1}} B_j & \text{if } j \leq i, \\
0 & \text{otherwise},
\end{cases}
\]
where $A_{j:i+1}$ refers to $A_j \cdots A_i$ when $i > j$ and $A_{i}$ when $i=j$. 
The matrix \(M\) is a specific type of semiseparable matrices, which is a rank-structured matrix where every submatrix contained on and below the diagonal has a rank of at most $N$, corresponding to the SSM’s state dimension.
% meaning its submatrices possess low-rank properties. 
Specifically, a semi-separable matrix \(M\) can be expressed as the sum of two components:
\[
M = UV^\top + K,
\]
where \(U\) and \(V\) capture the structured part, and \(K\) represents the lower-triangular portion. This structure ensures efficient computation with \(O(NL)\) complexity, which is significantly faster than the \(O(L^2)\) cost of standard Transformers-based models.

\noindent

As shown in Fig.~\ref{fig:detail_architect}, the \JambaDNA{} Mamba2 block is a scalable model architecture designed for efficient processing of input sequences. 
% Before the SSD layer, it integrates grouped-value projections and lightweight convolutional operations. 
It incorporates grouped-value projections and lightweight convolutional operations before the SSD layer.
These projections, along with 1D convolutions, facilitate flexible feature extraction and dimensionality reduction while maintaining computational efficiency. 
All data-dependent projections are computed in parallel at the start of the block, utilizing tensor parallelism to maximize the use of matrix multiplication units on modern GPUs.
Additionally, the Mamba2 blocks employ RMSNorm normalization~\citep{zhang2019root} both before input projection and after the SSD layer, enhancing training stability, particularly at large model scales. 
For the \JambaDNA{} Transformers block, we use a standard Transformers Decoder block as described in Section~\ref{sec:attention}.

% \textbf{\JambaDNA{}} builds upon the Mamba2 block and the Transformers-decoder block components. Referencing on insights from ~\citep{waleffe2024empirical} in natural language domains, \JambaDNA{} alternates Mamba2 and Transformer blocks in a 7:1 ratio. 
% This configuration has been demonstrated to effectively balance the strengths of both block types, resulting in optimal performance.

In the \JambaDNA{} architecture, the first block is a \JambaDNA{} Mamba2 block, which eliminates the need for explicit positional embeddings or mechanisms such as RoPE~\citep{su2024roformer}.
This design choice results in a \JambaDNA{} design that completely omits positional encoding. 
Furthermore, unlike the Jamba model~\citep{lieber2024jambahybridtransformermambalanguage}, the MLP layers in \JambaDNA{} do not use Mixture-of-Experts (MoE) configurations due to instability observed during fine-tuning for DNA-related downstream tasks. 
By leveraging this hybrid architecture, \JambaDNA{} excels in both short- and long-range tasks while enabling the robust generation of synthetic DNA sequences.

\subsection{Pretraining on Multi-Species Genomes}

\textbf{Dataset}
We pretrain \JambaDNA{} on a large-scale, multi-species genome dataset using next nucleotide (token) prediction (NTP). 
This dataset was curated from the Nucleotide Transformer~\citep{dalla2023nucleotide} and NCBI,
% The resulting collection of genomes was downsampled to include a total of 845 species, collectively comprising 160 billion nucleotides.
and was down-sampled to include 845 species, collectively comprising 160 billion nucleotides. 
Table~\ref{table:pretrain_data} provides a summary of the contribution of each genome class, represented as the number of nucleotides relative to the total nucleotide count in the dataset.
% The contribution of each class, in terms of the number of nucleotides relative to the total nucleotide count in the dataset, is summarized in Table~\ref{table:pretrain_data}.
A diverse and comprehensive genome dataset encompassing multiple species is essential for enabling the model to effectively interpret a wide range of genomic sequences. 
Such a dataset ensures that the model captures patterns and interactions representative of various biological systems, thereby enhancing its ability to generalize across species and tasks. 
% A summary of the training and validation datasets used during the pertaining stage is presented in Table~\ref{table:pretrain_data}.
% The multi-species dataset is collected from 
% Multi-species genomic data equips \JambaDNA{} to identify both shared and unique characteristics of genomic sequences across diverse organisms. 
% This diversity is particularly advantageous for deciphering complex genomic phenomena such as conserved regulatory elements, species-specific variations, and evolutionary adaptations. 

% This design avoids the pitfalls of k-mer tokenization, such as information leakage and computational inefficiencies, while retaining the fine-grained resolution necessary to capture subtle variations in genomic sequences. By modeling at the single-nucleotide level, \JambaDNA can accurately reflect the biological significance of minor sequence changes. 

\begin{table}[h]  
\centering  
\resizebox{\textwidth}{!}{  
\begin{tabular}{lcccc} 
\toprule  
\textbf{Class} & \textbf{\# Species (train)} & \textbf{\# Nucleotides (train)} & \textbf{\# Species (valid)} & \textbf{\# Nucleotides (valid)} \\  
\midrule  
Bacteria             & 647   & 16.5B   & 20  & 0.5B   \\ 
Fungi                & 44    & 2.0B    & 3   & 0.2B   \\  
Invertebrate         & 37    & 19.9B   & 2   & 1.9B   \\  
Protozoa             & 9     & 0.45B   & 1   & 0.05B  \\  
Mammalian Vertebrate & 28    & 65.2B   & 3   & 4.6B   \\  
Other Vertebrate     & 51    & 57.4B   & 6   & 6.0B   \\  
\midrule  
\textbf{Total}       & \textbf{845} & \textbf{160.75B} & \textbf{35} & \textbf{13.25B} \\
\bottomrule  
\end{tabular}}  
\caption{Statistics of multi-species pretraining data used in \JambaDNA{}}  
\label{table:pretrain_data}  
\end{table}

\textbf{Tokenizer}~ \JambaDNA{} employs a straightforward and effective base-level tokenization strategy, encoding each nucleotide (A, C, T, G) as an individual token.
% Base-level tokenization ensures that the model processes genomic data with high fidelity to its natural structure, allowing for nuanced interpretation and feature extraction. Unlike higher-order tokenization schemes, which often aggregate multiple bases into a single token, the base-level approach treats each nucleotide as a fundamental unit, preserving its individual contribution to genomic patterns. This tokenization method is particularly advantageous for capturing low-level sequence variations that can have significant biological implications, such as single nucleotide polymorphisms (SNPs) and point mutations.
This strategy ensures that the model processes genomic data with high fidelity to its natural structure, enabling nuanced interpretation and feature extraction. 
Unlike higher-order tokenization schemes that aggregate multiple bases into a single token, the base-level strategy treats each nucleotide as a fundamental unit, preserving its unique contribution to genomic patterns. 
This method is particularly advantageous for capturing low-level sequence variations with significant biological implications, such as single nucleotide polymorphisms (SNPs) and point mutations.

Previous transformer-based models like DNABERT2~\citep{zhou2023dnabert2} and Nucleotide Transformer~\citep{dalla2023nucleotide} faced challenges when utilizing base-level tokenization due to the resulting longer sequence lengths, which lead to higher computational costs and memory demands. 
\JambaDNA{} overcomes these limitations through its hybrid architecture. The Mamba2 blocks enable efficient processing of long sequences, allowing \JambaDNA{} to harness the fine-grained detail of base-level tokenization without compromising performance or scalability. This capability is particularly critical for tasks that require modeling extensive genomic contexts, such as enhancer-promoter interactions and chromatin state analysis.
% However, the hybrid Mamba2 and attention-based architecture of \JambaDNA{} effectively mitigates these issues. The Mamba2 blocks enable efficient processing of long sequences, allowing \JambaDNA{} to leverage the fine-grained detail of base-level tokenization without compromising performance or scalability. This capability is critical for tasks requiring the modeling of extensive genomic contexts, such as enhancer-promoter interactions and chromatin state analysis.

% \textbf{DNA Sequence Length Warm-up}~To enable \JambaDNA{} to generalize effectively across extensive genomic contexts, we use a multi-stage long-range warm-up procedure for our model. Initially, the model is trained with an 8,192-token context length to establish a foundation for capturing intermediate sequence dependencies. Subsequently, the context length is progressively extended to 32,768 and then 131,072 tokens, utilizing an additional 2\% of the original training steps. This staged approach allows the model to adapt to increasingly long dependencies while preserving earlier learned representations. Leveraging the hybrid Mamba2 and attention-based architecture, this strategy enables efficient processing of large-scale genomic spans, making HybriDNA adept at handling tasks that require long-range understanding.
\textbf{DNA Sequence Length Warm-up}~To enhance \JambaDNA{}'s ability to generalize effectively across longer genomic ranges, we implement a multi-stage warm-up procedure during the pretraining phase. The pretraining process begins by training the model with an 8,192 token context length, establishing a strong foundation for capturing intermediate sequence dependencies.
After that, the context length is gradually increased—first to 32,768 tokens and then to 131,072 tokens—with each extension undergoing additional training equal to 2\% of the training steps originally used for the 8,192 token context length.
This gradual extension enables the model to adapt to increasingly long-range dependencies and ensure efficient processing of large-scale genomic spans, equipping \JambaDNA{} to excel in tasks that demand long-range comprehension.

\subsection{\textcolor{black}{Downstream Fine-tuning}} 

\subsubsection{Discriminative Fine-tuning for DNA Understanding tasks}

To develop a DNA foundation model capable of handling both generative and understanding downstream tasks, \JambaDNA{} employs a GPT-like decoder-only architecture. However, a key limitation of autoregressive models, compared to bidirectional models, is their inability to incorporate information from future tokens into the embeddings of current tokens.
To address this issue, \JambaDNA{} introduces a novel \textit{echo embedding} technique during the fine-tuning stage for understanding tasks, drawing inspiration from the work of~\citep{springer2024repetitionimproveslanguagemodel}.

% inspired by the work of~\citep{springer2024repetitionimproveslanguagemodel}, \JambaDNA{} uses a novel echo embedding technique during the fine-tuning stage for understanding tasks. 

% For input sequence $x \in \mathbb{R}^{T}$, 
% we form its ``echo'' input by concatenating:
% \[
%   \mathrm{concat}[x, x].
% \]
% The hidden embeddings are then extracted from the final hidden layer, focusing on the last $T$ positions as below:
% as shown below:
% \[
%   \mathrm{embed}_{\theta}(\mathrm{concat}[x, x])
%   \bigl[T+1 : 2T\bigr].
% \]
% Finally we apply a mean-pooling 
% operation across all $T$ of those token embeddings to obtain $h_{\theta}(x^{(i)})$. 
% This ensures that the embedding $h_{\theta}(x^{(i)})$ incorporates contextual information 
% from subsequent tokens (the repeated second $x^{(i)}$).
% Next, we feed $h_{\theta}(x^{(i)})$ into a classification head 
% (e.g.,\ a linear layer with weights $W \in \mathbb{R}^{K \times d}$ and 
% bias $b \in \mathbb{R}^K$) to obtain the predicted probability distribution over the $K$ classes:
The core idea of this method is that repeating sequences facilitates the encoding of contextual information from subsequent elements into the embeddings. 
To illustrate, consider an input sequence $x$ and its corresponding label $y$ in a classification task with $K$ classes. 
For instance, given the input sequence $x = \mathtt{AACG}$, an ``echo" input is created by duplicating $x$: $x_{\text{echo}} = \mathtt{AACGAACG}$.
Hidden embeddings are then extracted from the final hidden layer, with a particular focus on the embeddings from the latter half of the sequence.
A mean-pooling operation is applied over these selected token embeddings to produce $h_{\theta}(x_{\text{echo}})$, which is designed to capture contextual information from the repeated segment of the input. 
This pooled vector, $h_{\theta}(x_{\text{echo}})$ is subsequently passed into a classification head,  typically consisting of a linear layer with weights $W \in \mathbb{R}^{d \times K}$ and bias $b \in \mathbb{R}^K$, to generate the predicted probability distribution over the $K$ classes:
\begin{equation}
    P(y|x) = \mathrm{softmax}(h_{\theta}(x_{\text{echo}})W + b).
\end{equation}
To optimize the model, the standard cross-entropy loss is employed to adjust the parameters of either the classification head alone ($W$ and $b$) or the entire model ($\theta$, $W$, and $b$). 
By incorporating bidirectional context into the autoregressive model, echo embeddings bridge the gap between traditional autoregressive embeddings and the complex demands of high-resolution genomic tasks, offering significant advantages for analyzing large genomic sequences.

% repetition enables embeddings to encode 
% contextual information from subsequent tokens.
% Specifically, let $x$ represents a input sequence, and $y$ denotes its label for a $K$-class 
% classification task. 
% For instance, if input sequence $x = \mathtt{AACG}$, we form its ``echo'' input by repeating $x$ twice: $x_{echo} = \mathtt{AACGAACG}$.
% The hidden embeddings are then extracted from the final hidden layer, specifically focusing on the last half positions.
% Next, a mean-pooling operation is applied across all the token embeddings to produce $h_{\theta}(x)$, ensuring that $h_{\theta}(x)$ incorporates contextual information from the repeated portion of the input (the second $x$).
% $h_{\theta}(x)$ is then passed through a classification head—such as a linear layer with weights $W \in \mathbb{R}^{d \times K}$ and bias $b \in \mathbb{R}^K$—to generate the predicted probability distribution over the $K$ classes:
% \[
%   p_{\theta, W, b}(y|x) = \mathrm{softmax}(h_{\theta}(x)W \;+\; b).
% \]
% The standard cross-entropy loss is used to optimize the parameters of the classification head (i.e., $W$ and $b$) or the parameters of the entire model (i.e., $\theta$, $W$, and $b$). 
% By incorporating bidirectional context into the autoregressive framework, echo embeddings effectively bridge the gap between traditional autoregressive embeddings and the requirements of high-fidelity genomic tasks. 
% This approach proves especially advantageous for tasks that demand a nuanced understanding of long genomic sequences.

A potential limitation of using echo embeddings for discriminative fine-tuning is the increased computational cost, as doubling the input length may raise memory requirements. 
However, \JambaDNA’s efficient hybrid architecture helps mitigate this burden, making the technique practical and scalable for a wide range of genomic analyses and applications.

% We apply mean pooling to produce an ``echo embedding'' $h_{\text{echo}}(x^{(i)})$:
% \[
%   h_{\text{echo}}\bigl(x^{(i)}\bigr) \;=\; 
%     \text{mean}\bigl(
%       \text{embed}\bigl(\theta; \text{concat}[\,x^{(i)}, x^{(i)}]\bigr)[T+1 : 2T]
%     \bigr).
% \]

% Here, we first concatenate the sequence $x$ with itself, i.e.\ 
% $\text{concat}[x, x] \in \mathbb{R}^{2T}$. The embedding function 
% $\text{embed}(\theta; \cdot)$ produces hidden representations for each of the 
% $2T$ positions. Afterward, we \emph{slice} (or select) the latter $T$ embeddings 
% that correspond to the ``second'' $x$, and 

% Let $\hat{p}(y \,\mid\, x^{(i)}) = \text{softmax}\bigl(\text{logits}(x^{(i)})\bigr)$ 
% denote the predicted probability distribution over the $K$ classes. 
% Then the \emph{echo-embedding classification loss} for a batch 
% $\{(x^{(i)}, y^{(i)})\}_{i=1}^N$ is simply the cross-entropy:
% \[
%   \mathcal{L}_{\text{echo}}(\theta) 
%   \;=\;
%   - \frac{1}{N} \sum_{i=1}^{N} 
%   \log 
%   \hat{p}\bigl(y^{(i)} \,\mid\, x^{(i)}; \theta\bigr),
% \]
% where $\theta$ now includes both the embedding/encoder parameters and 
% the classification head parameters $(W, b)$.

\subsubsection{Generative Fine-tuning for DNA Generation Tasks}\label{method:generative_finetune}
Autoregressive natural language models, such as ChatGPT, are capable of generating highly realistic, human-like text while adhering to human instructions to produce satisfactory responses~\citep{schulman2022chatgpt}.
In a similar vein, \JambaDNA{} is an autoregressive model pretrained on
multi-species genomic data at single-nucleotide resolution, unlocking the potential to design novel and realistic DNA sequences for a broad range of real-world applications.

To achieve this, we introduce a set of prompt tokens specifically designed to encode task-specific instructions. 
These prompt tokens are incorporated into the existing nucleotide vocabulary and initialized randomly within the embedding layer, which is expanded to include additional rows corresponding to the newly introduced token IDs.
\JambaDNA{} then predicts each nucleotide token $x_t$ autoregressively, conditioned on all preceding prompt tokens that specify the task-specific requirements.

Specifically, we optimize the
\emph{next token prediction} loss:
\begin{equation}
  \label{eq:causal-lm-loss}
  \mathcal{L}_{\text{generative}}(\theta)
  \;=\;
  -\,\frac{1}{T} \sum_{t=1}^{T}
     \log \Bigl(
       p_\theta\bigl(
         x_t
         \,\big|\,
         z_0, \ldots, z_{k-1},\;
         x_0, \ldots, x_{t-1}
       \bigr)
     \Bigr),
\end{equation}
where $\theta$ represents the complete set of model parameters, $z_{k-1}$ denotes the $k$-th prompt tokens, and $x_{t-1}$ represents the $t$-th generated nucleotide token.
By minimizing Eqn.~\ref{eq:causal-lm-loss}, \JambaDNA{} is trained to interpret the specialized tokens and generate realistic genomic sequences that align with specific design objectives.

% More concretely, let $\mathbf{E} \in \mathbb{R}^{V \times d}$
% be the original embedding matrix for a vocabulary of size $V$ and embedding dimension $d$.
% When we add $U$ new tokens, we form an augmented embedding matrix $\mathbf{E}' \in \mathbb{R}^{(V+U) \times d}$:
% \[
%   \mathbf{E}' \;=\; \begin{bmatrix}
%     \mathbf{E} \\
%     \mathbf{E}_{\text{new}}
%   \end{bmatrix},
% \]
% where $\mathbf{E}_{\text{new}} \in \mathbb{R}^{U \times d}$ is randomly initialized.

\section{Experiments}

In the experiment section, we aim to answer the following questions regarding the \JambaDNA{} models and their key capabilities: 
(1) How do the pretraining losses of \JambaDNA{} models compare across different scales and configurations? (2) Can the models achieve state-of-the-art performance on short-range understanding benchmarks, and how does scaling affect their effectiveness? 
(3) How do the models perform on long-range understanding tasks, and do they exhibit improved results with increased pretraining context length?
(4) Can the models generate realistic and desirable regulatory sequences across multiple species?
(5) How do the models compare to pure transformer-based models in terms of computational efficiency during training?

We benchmark our model against a series of recently proposed tasks, including DNABERT2~\citep{zhou2023dnabert2}, BEND~\citep{marin2024bend}, and Genomics LRB~\citep{poli2023genomics}. In selecting tasks for comparison, we adhere to the principle of encompassing a diverse range of challenges, encompassing both short and long-range capabilities. Additionally, we prioritize tasks that are biologically meaningful, covering a variety of species and functionalities within DNA-related areas.

\subsection{Pretraining curves}
\label{sec:pretraining}
\paragraph{Configurations}

We train three variants of \JambaDNA{} with 300M, 3B, and 7B parameters.
These models differ in the number of layers, hidden size, and learning rates used during training.
Despite these variations, all models share a consistent pretraining strategy. 
Instead of the commonly used masked language modeling (MLM) loss in genomics foundation models, we use a next-token-prediction (NTP) loss to enable generative capability. 
Our models are trained on NVIDIA A100/H100 and AMD MI300X GPUs. 
Details of the model architecture and pertaining configurations can be found in Appendix~\ref{appendix:model}.

During the pretraining stage, our 300M, 3B and 7B parameter models are trained on 0.5M tokens per batch, optimized for efficient utilization of computational resources and consistent training dynamics. Initially, the models are pretrained on sequences with a context length of 8,192 for 500k steps, resulting in a total of 250B tokens (approximately 1.5 epochs) in the first pretraining stage.
Following this, the models undergo further pretraining to extend their capabilities to handle larger context lengths. 
This two-stage pretraining strategy allows the models to gradually adapt to more complex and computationally demanding settings, ensuring robust performance across varying sequence lengths.

\begin{figure}[t]
\centering
\includegraphics[width=0.9\textwidth]{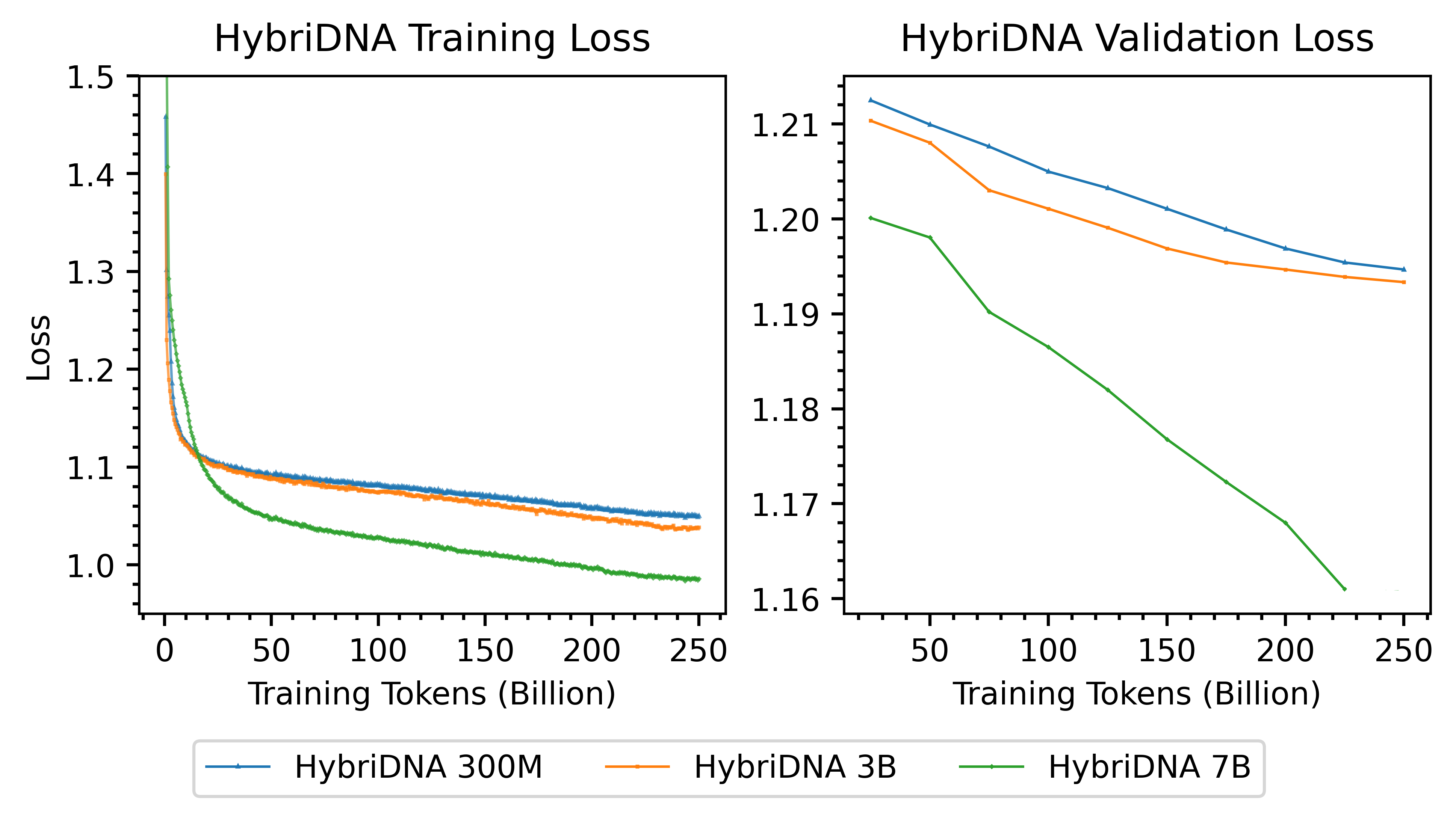}
\caption{Pretraining loss curves for \JambaDNA-300M, 3B, and 7B  models}
\label{fig:param_scaling_law}
\end{figure}

\paragraph{Scaling Behaviors} To investigate scaling law behavior, 
we analyze the training and validation losses during the pretraining stage for the 300M, 3B, and 7B models. 
As the model size increases, we observe consistent improvements in both training and validation losses, highlighting the advantages of larger models in capturing intricate genomic patterns. Detailed loss curves for each model variant are presented in Fig.~\ref{fig:param_scaling_law}, demonstrating the impact of scaling parameter sizes on pretraining model quality. These findings align with theoretical expectations of scaling laws in deep learning and genomics-specific modeling.

We further demonstrate the effectiveness of the hybrid architecture compared to the pure Mamba2 architecture. The training and validation losses for pretraining two comparable models, each with 300 million parameters, are shown in Fig.~\ref{fig:Mamba_Jamba_Comp}.

% missing a graph comparing Jamba/Mamba here, Mamba2 model is still under pretraining. ~2 days remaining.

% For longer-context further pretraining, we present the loss curve of 300M model in training with context length 8192, 32768 and finally 131,072 tokens in Fig. \ref{fig:context_scaling_law}. 

\subsection{Baselines}
We compare \JambaDNA{} with 5 state-of-the-art DNA foundation models:  NT-500M-human, NT-2.5B-MS, DNABERT-2, HyenaDNA-medium-160k and Caduceus-Ph-131k.

\textbf{NT-500M-human} is a Transformer-encoder based model with 500M parameters pretrained on GRCh38/hg38 human reference genome, which contains approximately 3.2B nucleotide bases. It employs a k-mer tokenization method with $k$ set to $6$ and is pretrained using the standard Masked Language Modeling (MLM) objective. The pretraining context length is 1,000.

\textbf{NT-2.5B-MS} is another variant in the Nucleotide Transformer series, featuring a larger size of 2.5B parameters.
It was pretrained on a multi-species dataset comprising 174 billion nucleotides from a total of 850 species. The other pretraining details are consistent with those of NT-500M-human.

\textbf{DNABERT-2} is also a Transformer-encoder based model with 112M parameter size. It is pretrained on a multi-species dataset containing approximately 32.5 billion nucleotide bases from 135 species. It improves the tokenizer with the Byte Pair Encoding (BPE) method and is also trained with a standard MLM objective with a context length of 512.

\textbf{Caduceus-Ph-131k} builds upon the Mamba1 architecture, which employs selective state space models for long-range sequence processing. It enables Bi-directional sequence modeling using BiMamba block. The model is trained on GRCh37/hg37 human reference genome with about 3.2B nucleotide bases using MLM objective. This variant has 7.73M parameter size and uses nucleotide-level tokenization with a pretraining context length of 131,072.

\textbf{HyenaDNA-Medium-160k} utilizes the Hyena operator, derived from state space models, for computationally efficient long-range sequence modeling. It is pretrained on GRCh38/hg38 human reference genome using next-token prediction objective. This specific variant has 14.2M parameters and uses nucleotide-level tokenization with a pretraining context length of 160,000.

For all baseline models, we utilize the pretrained weights provided in their respective original codebase. Since our evaluation is mainly focused on eukaryote-related tasks, Evo is excluded from the comparison.

\subsection{Short-range understanding benchmark (GUE, BEND)}
\label{sec:downstream}

\textbf{Genome Understanding Evaluation (GUE)}~\citep{zhou2023dnabert2} aggregates 28 datasets across 9 tasks, encompassing input lengths from 70 to 512 base pairs. GUE serves as a standardized evaluation suite, measuring the effectiveness of genomic foundation models in multi-species genome classification. The GUE benchmark assesses a model’s ability to analyze short-range genomic sequences across multiple biologically significant tasks. Promoter Detection (PD) and Core Promoter Detection (CPD) identify regulatory regions that initiate transcription, crucial for understanding gene expression control. Splice Site Detection (SS) predicts locations where pre-mRNA is spliced, affecting transcript diversity and protein function. Transcription Factor Binding Site (TFBS) Prediction determines whether a sequence contains motifs for transcription factors, which regulate gene activity. Epigenetic Marks Prediction (EMP) involves predicting histone modifications and DNA methylation, key regulators of chromatin state and gene expression. Lastly, COVID Variant Classification (CV) tracks viral genome mutations, aiding epidemiological surveillance. Collectively, these tasks provide a comprehensive measure of a model’s capacity to decode genomic structure and function.

\noindent\textbf{BEND}~\citep{marin2024bend} evaluates models on a collection of realistic and biologically meaningful tasks defined on the human genome. It emphasizes the importance of capturing intricate genomic data features through tasks that are comprehensive and provide a standardized evaluation methodology for genomics foundation models. 
For evaluation, we select the three largest short-range tasks across 3 different datasets from the benchmark.
Chromatin Accessibility Prediction assesses whether DNA is in an open or closed state, influencing gene expression potential. Histone Modification Prediction determines chemical changes to histones that affect chromatin structure and transcriptional activity. CpG Methylation Prediction identifies DNA methylation patterns at CpG sites, which play a critical role in gene silencing and disease progression. These tasks are essential for understanding the regulatory landscape of the genome and its implications for cellular function and disease mechanisms.

\subsubsection{GUE}
\textbf{Settings}
DNABERT-2 \cite{zhou2023dnabert2} introduces the GUE benchmark, which includes a series of short-range classification tasks across multiple species, such as Promoter Detection (PD), Core Promoter Detection (CPD), Splice Site Detection (SS), Transcription Factor Prediction (TF) for human and yeast, Epigenetic Marks Prediction (EMP) and Covid Variant Classification (CV). 
Following the same setting as DNABERT-2, we use metrics of Matthews Correlation Coefficient (MCC) for all tasks,  except for the Covid task, where we use the F-1 score according to the GUE dataset's original setting. 
The hyperparameters, training epochs, and evaluation strategies follow exactly from the original paper. We fine-tune all model parameters and use the hidden state of the last token for embedding representation for decoder-only models. Training epochs and evaluation steps are also consistent with the original paper. We apply a learning rate of $5e-5$ for our 300M model, $3e-5$ for our 3B model, and $1e-5$ for our 7B model across all tasks.

\noindent\textbf{Results} We take the mean MCC/F1-score value of tasks in the same category in the following table.  For detailed results on each task, refer to Appendix \ref{appendix:benchmark_gue}.
The suffix ``(E)'' in the model name within the table indicates that echo embedding was applied during discriminative fine-tuning. Results of our model are presented in Table~\ref{tab:gue}.

\begin{table}[t]
\centering
\resizebox{\textwidth}{!}{%
\begin{tabular}{llccccccc}
\toprule
\textbf{Type} & \textbf{Model} & \textbf{PD(H)} & \textbf{CPD(H)} & \textbf{SS(H)} & \textbf{TF(H)} & \textbf{TF(M)} & \textbf{EMP(Y)} & \textbf{CV(V)}\\
& &\emph{(MCC)} & \emph{(MCC)} & \emph{(MCC)} & \emph{(MCC)} & \emph{(MCC)} & \emph{(MCC)} & \emph{(F-1)} \\
\midrule
\multirow{3}{*}{\textbf{Encoder}} & DNABERT-2 &83.96 &71.81  & 85.42 &68.71& 70.00 & 55.98&71.02 \\
& NT-2.5B-MS & \textbf{88.15} & 71.57 & 89.35 & 63.21&67.02 & 57.64& 73.04\\
& NT-500M-human & 82.96 & 66.79 &78.63  & 61.92& 45.24 & 45.35 &50.82 \\
& Caduceus-Ph & 82.36 & 67.03 &71.80  &65.17 &62.28 & 51.05&40.35 \\
\midrule
\textbf{Decoder} & HyenaDNA & 80.14 &69.22  & 77.76 &61.74 &64.39 & 47.15&25.88 \\
\midrule
\multirow{6}{*}{\textbf{Our}} & \JambaDNA-300M &83.29 & 68.87&87.74 &68.37 &75.32 &67.38 &73.81 \\
& \JambaDNA-300M (E) &83.67 & 69.96& 88.72&69.70& 75.73& 68.25 & 73.90 \\
& \JambaDNA-3B &85.40 &69.50 &89.01 &70.48 &75.43 & \textbf{69.06} & 74.05\\
& \JambaDNA-3B (E) & 85.55 &70.71 &89.10 & 71.13&77.14 & 68.97& \textbf{74.88}\\
& \JambaDNA-7B & 86.53&71.37 &90.09 &70.72 &78.02 & 63.05&74.02\\
& \JambaDNA-7B (E) & 88.10 & \textbf{72.03} & \textbf{90.12} & \textbf{72.01} & \textbf{79.02} & 65.30 & 74.30 \\
\bottomrule
\end{tabular}%
}
\caption{Results on the GUE Benchmark, which encompass a series of short-range classification tasks across multiple species, including Promoter Detection (PD), Core Promoter Detection (CPD), Splice Site Detection (SS), Transcription Factor Prediction (TF), Epigenetic Marks Prediction (EMP) and Covid Variant Classification (CV). 
The suffix ``(H)'' denotes the human genome, ``(M)'' the mouse genome, ``(Y)'' the yeast genome, and ``(V)'' the virus genome.
Additionally, the suffix “(E)” in the model name indicates that echo embedding was applied during discriminative fine-tuning.
}
\label{tab:gue}
\end{table}
\subsubsection{BEND}

\textbf{Settings} BEND paper \cite{marin2024bend} presents a series of biologically meaningful tasks for genomics foundation model derived from a series of studies. We select the three short-range supervised tasks: Chromatin Accessibility, Histone Modification, and CpG Methylation. For the Histone Modification Tasks, the training process follows the original paper. We freeze the embedding of the models and train a downstream CNN model for 100 epochs. For autoregressive models like HyenaDNA and our \JambaDNA{} model, we use the mean of the hidden state of the sequence as embedding representations. The model with the lowest validation loss is tested and the metric reported is the mean AUROC score.

\noindent\textbf{Results} We directly report the AUROC score of each model on the three tasks in Table~\ref{tab:bend}.

\begin{table}[t]
\centering
\resizebox{\textwidth}{!}{%
\begin{tabular}{llccc}
\toprule
\textbf{Model Type} & \textbf{Model} & \textbf{Chromatin} & \textbf{Histone} & \textbf{CpG} \\
& & \textbf{Accessibility} & \textbf{Modification} & \textbf{Methylation} \\
& & \emph{(AUROC)} & \emph{(AUROC)} & \emph{(AUROC)} \\
\midrule
\multirow{3}{*}{\textbf{Encoder Models}} & DNABERT-2 & 0.81 & 0.79 & 0.90 \\
 & NT-2.5B-MS & 0.79 & 0.78 & 0.92 \\
 & NT-500M-human & 0.74 & 0.76 & 0.88 \\
 & Caduceus-Ph & 0.83 & 0.77 & 0.91 \\
 
\midrule
\textbf{Decoder Models} & HyenaDNA & 0.81 & 0.77 & 0.87 \\
\midrule
\multirow{3}{*}{\textbf{Our Model}}& \JambaDNA-300M & 0.78 & 0.77  & 0.88  \\
& \JambaDNA-3B& 0.82 &  0.79  & 0.92  \\
& \JambaDNA-7B& \textbf{0.84}  & \textbf{0.79} & \textbf{0.93}  \\
\bottomrule
\end{tabular}%
}
\caption{Results on the BEND Benchmark, which includes Chromatin Accessibility, Histone Modification, and CpG Methylation tasks. 
}
\label{tab:bend}
\end{table}

\subsection{Genomics long-range benchmark (LRB)}

The Genomics Long-Range Benchmark (LRB)~\cite{poli2023genomics} is designed to evaluate tasks that require understanding long-range context within genomic sequences. 
To assess models’ ability to capture dependencies across extended genomic regions, 
we select two representative tasks across distinct datasets that inherently demand long-range sequence comprehension:
Causal eQTL Variant Effect Prediction and OMIM Variant Effect Prediction. 
The Causal eQTL Variant Effect Prediction task evaluates the impact of genetic variants on gene expression levels, linking non-coding mutations to functional changes. 
The OMIM Variant Effect Prediction task focuses on identifying pathogenic mutations associated with Mendelian diseases, aiding in genetic diagnostics and precision medicine. 
These tasks test a model’s proficiency in analyzing complex gene regulation and variant interpretation over extended genomic regions.

During discriminative fine-tuning, all models were trained using frozen embeddings generated in the same way as those used in the BEND benchmark. These embeddings were passed through an MLP classifier, with all models sharing identical architectures and hyperparameters to maintain consistency.
We report accuracy and AUROC metrics for all tasks, with detailed results summarized in Table~\ref{tab:lrb}.

\begin{table}[t]
\centering
\resizebox{\textwidth}{!}{%
\begin{tabular}{llccccc}
\toprule
\textbf{Model Type} & \textbf{Model} & \multicolumn{2}{c}{\textbf{Causal eQTL}} & \textbf{OMIM} \\
\cmidrule(lr){3-4} \cmidrule(lr){5-5} \cmidrule(lr){6-6} \cmidrule(lr){7-7}
 & & \textit{Fine-tune} & \textit{Zero-shot} & \textit{Zero-shot} \\
 & & \emph{(AUROC)} & \emph{(AUROC)} & \emph{(AUPRC)} \\
\midrule
\multirow{3}{*}{\textbf{Encoder Models}} & DNABERT-2 & 0.72 & 0.50 & 0.002 \\
 & NT-500M-human & 0.72 & 0.51 & 0.003 \\
 & Caduceus-Ph & 0.68 & 0.49 & 0.002 \\
\midrule
\textbf{Decoder Models} & HyenaDNA & 0.71 & 0.51 & 0.002 \\
\midrule
\multirow{3}{*}{\textbf{Our Model}} & \JambaDNA-300M (8k) & 0.71 & 0.51 & 0.003 \\
 & \JambaDNA-300M (32k) & 0.72 & 0.51 & 0.003  \\
 & \JambaDNA-300M (131k) & \textbf{0.74} & \textbf{0.51} & \textbf{0.003}  \\
\bottomrule
\end{tabular}%
}
\caption{Results on the LRB Benchmark, which includes Causal eQTL Variant Effect prediction, OMIM Variant Effect prediction tasks.}
\label{tab:lrb}
\end{table}

\subsection{Designing realistic synthetic cis-regulatory elements (CREs)}
\label{sec:generation}

\textbf{regLM}~\citep{lal2024reglm} is a framework that combines  autoregressive language models with supervised sequence-to-function tasks to design synthetic cis-regulatory elements (CREs). 
This framework highlights the capability of models to generate regulatory sequences with specific desired properties. 
Specifically, (1) Human Enhancer Generation involves designing enhancers that drive gene expression in specific cell types, which is crucial for gene therapy and functional genomics. 
(2) Yeast Promoter Generation focuses on engineering promoters with defined transcriptional activity, supporting biotechnology applications such as industrial enzyme production. 
These tasks demonstrate a model’s capacity to generate biologically plausible DNA sequences that can be experimentally validated for targeted gene regulation.
 
We followed the experimental setup established by \textbf{regLM}. For comparison, we utilized HyenaDNA~\citep{poli2023hyenadna}, which is the default decoder-only model in the regLM study. Further details on the fine-tuning configurations can be found in Appendix~\ref{appendix:finetune_config}.

\subsubsection{Cell type-specific human enhancer generation}

The human enhancer generation task focuses on designing desired human enhancer genomic sequences for three specific cell line types: HepG2, K562, and SK-N-SH. 
Each sequence includes a three-digit label (ranging from 0 to 3) that represents the enhancer's activity strength in a given cell line. 
The model is fine-tuned on a dataset consisting of 670,000 training samples of 200bp enhancers with varying activity levels, utilizing the prompt tokens described in Section~\ref{method:generative_finetune}.

During evaluation, the model is tasked with generating 600 sequences for the specific labels\{300, 030, 003\}, which represent high enhancer activity in a particular cell line. 
These labels are extremely rare in the training data,  accounting for only 0.16\% of the total training set.
These generated sequences are evaluated using a separately trained scoring model from regLM to assess their actual enhancer activity in the respective cell types. 
Specifically, beam search decoding is used for all models during sequence generation to ensure a fair comparison.
The baseline for evaluation is the fine-tuned HyenaDNA model variant, "hyenadna-medium-160k-seqlen," as referenced in the original regLM paper.

The evaluation metrics for the generated sequences are defined as follows:
\begin{enumerate}
\item Top-1 activity: The highest predicted enhancer activity for each cell type.
\item Mean activity: The average of the top 100 predicted enhancer activity scores for each cell type.
\item Diversity: The mean of pair-wise edit distance of the top 100 predicted sequences across all cell types, measuring the overall diversity of high-quality generated sequences.
\end{enumerate}

The results are summarized in Table~\ref{tab:human_enhancer_comparison}. Result shows that our model outperforms both the held-out test set and HyenaDNA in generating higher-activity enhancer sequences for each cell line, while also maintaining greater diversity.

\begin{table}[t]
\centering
\resizebox{\textwidth}{!}{  
\begin{tabular}{lccccccc}
\toprule
\multirow{2}{*}{Model} & \multicolumn{2}{c}{HepG2} & \multicolumn{2}{c}{K562} & \multicolumn{2}{c}{SK-N-SH} & Diversity \\
\cmidrule(lr){2-3} \cmidrule(lr){4-5} \cmidrule(lr){6-7} \cmidrule(lr){8-8}
 & Top-1 & Mean &  Top-1 & Mean &  Top-1 & Mean & Mean Edit Distance \\
\midrule
Held-out Test  &  6.2     & 2.6      &    5.5      &  2.4           &    5.1   &  1.6     &   \textbf{110.10}     \\

HyenaDNA      &  5.5     & 4.0      &    4.3      &  3.8           &    5.2   &  2.3     &    98.50      \\
\JambaDNA-300M &  \textbf{7.3}     &  \textbf{5.4}     &    \textbf{7.8}      &  \textbf{6.2}            & \textbf{6.6}      &  \textbf{4.7}     &   108.74       \\
\bottomrule
\end{tabular}
}
\caption{Comparison between \JambaDNA-300M and HyenaDNA on the human enhancer generation task. Metrics include Top-1 activity, Mean activity, and Diversity for each cell line type (HepG2, K562, SK-N-SH).}
\label{tab:human_enhancer_comparison}
\end{table}

\subsubsection{Yeast promoter generation}
The yeast promoter generation task follows a similar setup to the human enhancer generation task. 
However, instead of three cell lines, this task utilizes a two-digit label representing promoter activity in complex and defined media, with activity levels ranging from 0 to 4.
Since HyenaDNA was pretrained primarily on human genomic data, the regLM study trained the HyenaDNA model from scratch using yeast promoter data. In contrast, our model’s pretraining corpus already includes multi-species data, including yeast genomes. 
As a result, rather than training our model from scratch, we fine-tune it from a pretrained checkpoint using the yeast promoter dataset.

The model is fine-tuned on a dataset comprising 7.4 million training samples of 80bp promoters with varying activity levels.
During evaluation, it is prompted to generate sequences with label \{40, 04\}, which represents only 0.34\% of the training samples. The evaluation steps and metrics are identical to those in the human enhancer generation task., The results are summarized in Table~\ref{tab:yeast_promoter_comparison}. 
Notably, our model achieves higher promoter activity scores across both media types and generates more diverse sequences compared to the HyenaDNA baseline.

\begin{table}[h]
\centering
\resizebox{\textwidth}{!}{  
\begin{tabular}{lccccc}
\toprule
\multirow{2}{*}{Model} & \multicolumn{2}{c}{Complex Media} & \multicolumn{2}{c}{Defined Media} & Diversity \\
\cmidrule(lr){2-3} \cmidrule(lr){4-5} \cmidrule(lr){6-6}
 & Top-1 & Mean & Top-1 & Mean &  Mean Edit Distance \\
\midrule
Held-out Test & 16.0 & 5.9 & 15.7 & 6.7 & 28.8 \\

HyenaDNA &  16.8     &  11.4     &  16.3     & 10.8      &    27.3      \\
\JambaDNA-300M &  \textbf{18.2} & \textbf{15.0}           &  \textbf{17.6}     &   \textbf{13.5}    &  \textbf{30.7}         \\
\bottomrule
\end{tabular}
}
\caption{Comparison between \JambaDNA-300M and HyenaDNA on the yeast promoter generation task. Metrics include Top-1 activity, Mean activity, and Diversity for both media types (Complex and Defined).}
\label{tab:yeast_promoter_comparison}
\end{table}

\subsection{Computational efficiency}

% To measure the computational efficiency of our HybriDNA model comparing to a standard Transformers model (with the same Transformers block we adopt in our hybrid model) of the similar parameter size, especially during the training, we compare them in these two following metrics:

To evaluate the computational efficiency of the \JambaDNA{} model compared to a standard Transformer model with a similar parameter size, particularly during the training phase, we use the following two metrics:

\begin{enumerate}
\item {Tokens/second per GPU: This metric assesses the throughput of a single GPU by measuring the number of tokens it can process each second during the pretraining stage. For each context length, the batch size is set to the maximum number to fit within the GPU memory. }
\item {GPU memory cost (GB): This metric measures the amount of GPU memory consumed when training models with a fixed context length and a batch size of 1.}
\end{enumerate}

We evaluate both models using four NVIDIA A100 GPUs (80G memory) with DeepSpeed Zero-1 Stage optimization and BF16 mixed-precision training. 
Both models comprise approximately 300M parameters. 
The Transformer model uses Flash Attention 2 optimization, while the Mamba2 layers in the \JambaDNA{} model are implemented using CUDA. We test the pretraining of both models at various context lengths, ranging from 2k tokens to 65k tokens, doubling the sequence length at each step.

\begin{figure}[t]
\centering
\includegraphics[width=0.9\textwidth]{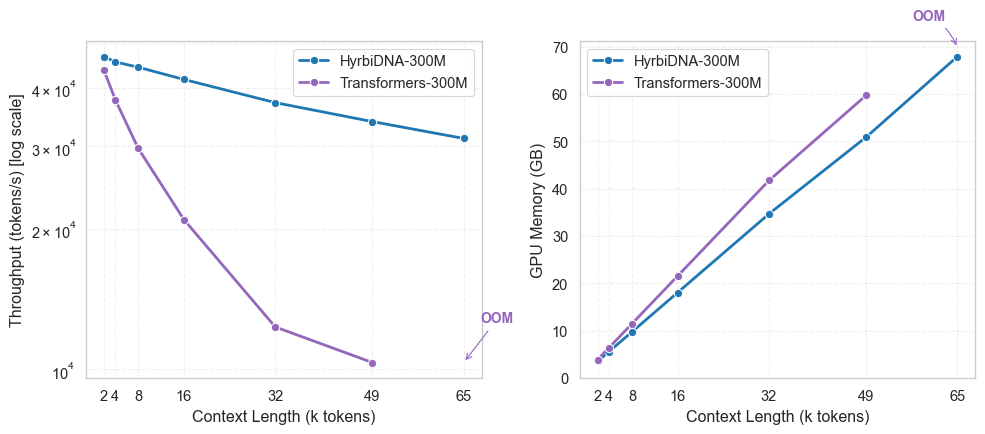}
\caption{
Comparison of training throughput and GPU memory consumption between \JambaDNA{} and a pure Transformer model with comparable parameters (e.g., 300M)
}
\label{fig:efficiency}
\end{figure}

As shown in Fig.~\ref{fig:efficiency}, the \JambaDNA{} model achieves significantly higher training throughput than standard Transformer models, especially when processing context lengths exceeding 32,000 tokens. For instance, at a context length of 49,000 tokens, the throughput of \JambaDNA{} is approximately 3.4 times higher than that of Transformers. This performance gap widens as context length increases, highlighting the superior efficiency of our model compared to Transformers.

In terms of GPU memory usage, \JambaDNA{} consistently demonstrates greater efficiency than Transformer models, even those optimized with advanced techniques such as Flash Attention 2~\citep{dao2023flashattention}. 
Notably, at context lengths around 65,000 tokens, a standard Transformer model encounters Out-Of-Memory (OOM) issues on A100 GPUs. 
These findings underscore the exceptional capability of our hybrid model to effectively manage larger context lengths, a crucial advantage for complex long-range DNA-related tasks.

% Regarding GPU memory usage, even though modern Transformer models are highly optimized with techniques such as Flash Attention 2, our HybriDNA consistently surpasses a standard Transformer model in efficiency. Moreover, for larger context lengths, such as 65k tokens, a standard Transformer model encounters Out-Of-Memory (OOM) errors in A100 GPUs and suffers from reduced throughput due to its inherent quadratic complexity. These results demonstrate the superior capability of our hybrid model to handle larger context lengths effectively, which is crucial for complex long-range DNA-related tasks.

\section{Related Work}

\subsection{DNA Foundation Models}

The advent of high-throughput sequencing technologies has produced vast amounts of genomic data, presenting an unprecedented opportunity for deep learning to uncover complex relationships and dependencies in DNA sequences. Recent advancements in genome language modeling have demonstrated their effectiveness across a wide range of downstream applications, including promoter prediction~\citep{le2022bert,zhang2022ipro}, gene expression  prediction~\citep{avsec2021effective}, DNA methylation prediction~\citep{jin2022idna}, chromatin state analysis~\citep{lee2022learning}, promoter-enhancer interaction prediction~\citep{chen2022capturing,ni2022epi}
TF-DNA binding prediction~\citep{wang2022towards}, variant effect prediction~\citep{rozowsky2023tex},
gene network prediction~\citep{theodoris2023transfer} and more.
More recently, inspired by advancements in natural language processing, researchers have begun developing DNA foundation models. These include, but are not limited to: (1) encoder-only models such as DNABERT, DNABERT-2, Nucleotide Transformer,  and Caduceus; and (2) decoder-only models such as HyenaDNA and Evo.

% \textbf{Encoder-only models}~
\textbf{DNABERT}~\citep{ji2021dnabert} is an early foundation model designed to interpret the human genome from a language perspective. By adapting the BERT framework with Transformers architecture, it captures a transferable understanding of human genome reference sequences. This single pretrained Transformer model achieves state-of-the-art performance in tasks such as predicting promoters, splice sites, and transcription factor binding sites, after fine-tuning on small task-specific labeled datasets. The model contains 86M parameters and operates with a context length of 512 on the hg38 human reference genome dataset.

\textbf{DNABERT-2}~\citep{zhou2023dnabert2} builds on its predecessor by employing Byte Pair Encoding (BPE) for tokenization, which improves computational efficiency and representation quality. It also incorporates Attention with Linear Biases (ALiBi) with Transformers-Encoder layers, enabling the model to process longer input sequences effectively. DNABERT-2 achieves state-of-the-art results on the Genome Understanding Evaluation (GUE) benchmark, showcasing its capacity to address diverse genomic tasks. The model consists of 112M parameters and is trained on a multi-species dataset comprising 135 species with a total of 32 billion nucleotides and a context length of 512.

\textbf{Nucleotide Transformer (NT)}~\citep{dalla2023nucleotide} is a scalable genomics foundation model, built on an encoder-only Transformer architecture, with parameter sizes ranging from 500M to 2,500M, based on encoder-only Transformer architecture. 
Its multi-species variant is pretrained on genomic data from 850 species, employing a non-overlapping k-mer tokenization method that effectively reduces tokenized sequence lengths. 
Additionally, two human-specific versions are trained separately on the hg38 human reference genome dataset and the 1000 Genomes Project. All pretraining is conducted with a context length of 1,000 tokens. 
% Two human-specific versions are trained separately on the hg38 human reference genome dataset and the 1000 Genomes Project. All pretraining uses a context length of 1,000 tokens. 
% The model's versatility lies in its ability to generalize nucleotide sequence representations across a wide range of genomic tasks, demonstrating the scalability potential of large models in genomics.

\textbf{Caduceus}~\citep{schiff2023caduceus} introduces the bi-directional Mamba1 architecture, specifically designed for DNA sequence modeling. By incorporating reverse complement (RC) equivariance at the architectural level, Caduceus is optimized for long-range DNA sequence modeling. The model effectively captures the intricate understanding required for DNA sequence tasks. The Caduceus series features parameter sizes ranging from 500K to 7M, with a context length of 131k, and is trained on the hg38 dataset.

\textbf{HyenaDNA}~\citep{poli2023hyenadna} utilizes the Hyena operator, a recurrence of gating and implicitly parametrized long convolutions, to handle long-range genomic sequences, enabling the processing of input contexts up to 1 million tokens with single-nucleotide resolution. 
This model shows effectiveness in tasks requiring long-range understanding, such as analyzing DNA fragments far apart, beyond the context window of traditional Transformer models. HyenaDNA is trained on the hg38 human reference genome dataset, with parameter sizes ranging from 1.7M to 50M and context lengths varying from 1k to 1M.

\textbf{Evo}~\citep{meier2023evo} is a 7-billion-parameter foundation model built on the StripedHyena architecture and trained on 2.7 million raw prokaryotic and phage genome sequences. 
It integrates multiple biological modalities, including DNA, RNA, and proteins. With a context length of 131k nucleotide bases, Evo delivers superior performance in sequence modeling and functional design tasks, spanning molecular to genome-scale applications.

Two concurrent works, \textbf{GenomeOcean} \cite{Zhou2025genomeocean} and \textbf{GENErator} \cite{wu2025generator}, have recently emerged in the field of DNA foundation models. GenomeOcean is a 4B-parameter model trained on diverse meta-genomics samples, optimizing for microbial species representation and achieving faster genome generation. GENErator, a 1.2B-parameter model with a 98k context length, is trained on 386B base pairs of eukaryotic DNA and excels in generating protein-coding sequences, designing promoter sequence and optimizing promoter activity. These models further expand genomic sequence modeling and generation capabilities.

\subsection{Hybrid Models in General Domains}
Recent advancements in Mamba-based hybrid models for NLP tasks combine the efficiency of SSMs with the expressiveness of attention mechanisms, excelling in long-context scenarios. Innovations include Jamba's~\citep{lieber2024jambahybridtransformermambalanguage} integration of Transformer, Mamba, and Mixture-of-Experts layers for sequences up to 256k tokens, 
Zamba's \citep{glorioso2024zambacompact7bssm} compact 7B model with shared self-attention for reduced latency, and SAMBA's~\citep{ren2024sambasimplehybridstate} sliding window attention for efficient handling of sequences up to 1M tokens. Other notable contributions include Taipan's~\citep{vannguyen2024taipanefficientexpressivestate} selective attention layers for scalability and Waleffe's~\citep{waleffe2024empirical} versatile 8B hybrid architecture combining Mamba2, self-attention, and MLP layers. These models achieve strong results across various short- and long-range benchmarks.

\section{Conclusion}
% \textcolor{red}{Guoqing}
\label{sec:conclusion}

In this work, we develop a class of decoder-only DNA language models built on a hybrid Transformer-Mamba2 architecture. 
This design harnesses the unique strengths of its two core components to enable efficient and precise modeling of DNA sequences.
By integrating Mamba2 layers, our model can process extremely long DNA sequences at single-nucleotide resolution with remarkable computational efficiency.
Pretrained on large-scale, multi-species genomes at single-nucleotide resolution with a next-token prediction objective, \JambaDNA{} demonstrates foundational capabilities in both understanding and designing genomic sequences.
Through echo embedding discriminative fine-tuning, \JambaDNA{} achieves state-of-the-art performance across 33 biologically significant DNA understanding tasks from
the BEND, GUE, and LRB benchmarks. 
Through generative fine-tuning, \JambaDNA{} exhibits remarkable proficiency in generating synthetic cis-regulatory elements with desirable functional properties.
These results highlight \JambaDNA’s versatility and establish its potential as a powerful foundation model for advancing DNA research and applications.

Looking ahead, there are several exciting directions to further explore. 
These include: (1) Expanding the pretraining dataset to include a greater number of nucleotide tokens and species classes, enabling broader generalization across downstream tasks involving diverse species. (2) Conducting more downstream fine-tuning tasks with diverse and significant scientific impacts, and performing wet-lab experiments to further validate the sequences designed by \JambaDNA.

\section{Acknowledgements}
We would like to thank Sumit Basu from the Health Futures team at Microsoft Research for his valuable insights on benchmarks and writing; Jingyun Bai for her design; and Ran Bi, Hannes Schulz, Jean Helie, and Maik Riechert for their engineering support. Additionally, we sincerely acknowledge the entire AI for Science team at Microsoft Research for their continuous support.

% In this work, we introduce a family of large-scale hybrid Mamba2 Transformers GFM. 

\newpage

\begin{appendices}
\section{Model Details}
\label{appendix:model}
\subsection{Model architecture}

The \textbf{\JambaDNA} model employs a hybrid Transformer-Mamba2 architecture. The architecture interleaves Transformer and Mamba2 layers in a 7:1 ratio. The Transformer layer is placed in the fourth of every eight layers. Our three model variants—300M, 3B, and 7B—differ in their hidden dimension size and layer configurations. The details of each model architecture are summarized in Table~\ref{table:model_architecture}.

\begin{table}[htbp]  
\centering  
\begin{tabular}{lccccc}  
\toprule  
\textbf{Model Variant} & \textbf{\# Layers} & \textbf{Hidden Size} & \textbf{Intermediate Size} & \textbf{\# Heads} & \textbf{Head Dim} \\ 
\midrule  
7B   & 32 & 4096 & 8192 & 128 & 64 \\
3B   & 16 & 4096 & 8192 & 128 & 64 \\
300M\footnotemark & 24 & 1024 & 2048 & 32 & 64 \\
\bottomrule  
\end{tabular}  
\caption{Model configurations for three variants of \JambaDNA{}}  
\label{table:model_architecture}  
\end{table}

\subsection{Pretraining configuration}

\footnotetext{You may notice that HybriDNA-300M model has 32 layers. We draw inspiration from the configuration of Jamba-1.5 model: \href{https://huggingface.co/ai21labs/AI21-Jamba-1.5-Large/blob/main/config.json}{https://huggingface.co/ai21labs/AI21-Jamba-1.5-Large/blob/main/config.json}.}

% Fig. \ref{fig:pipeline} is a detailed pipeline diagram of our \textbf{\JambaDNA{}} model, including its multi-stage training process, the detail illustration of hybrid architecture, and two types of downstream usage of our model. 
During the pretraining stage, we trained the \JambaDNA{} models using a next token prediction objective. 
The training utilized the Adam \cite{adam} optimizer with a learning rate schedule and standard exponential decay rates $\beta_1 = 0.9$, $\beta_2 = 0.95$, and $\epsilon = 1\mathrm{e}{-8}$. 
All model variants underwent a Warmup phase of 2,000 steps, with a total of 500,000 training steps. 
The learning rates were set at $1\mathrm{e}{-3}$ for the 300M model, $6\mathrm{e}{-4}$ for the 3B model, and $1\mathrm{e}{-4}$ for the 7B model. 
Mamba-based models demonstrate a higher tolerance for learning rates compared to standard Transformer architectures, highlighting their stability during optimization.

Our pertaining was conducted on the following hardware configurations: the 300M model on 8 AMD MI300X GPUs, the 3B model on 8 NVIDIA H100 GPUs, and the 7B model on 64 AMD MI300X GPUs. 
Models are trained for approximately 300 hours for the 300M and 7B variants, and 500 hours for the 3B model. 
These configurations ensured efficient utilization of computational resources and stable training for large-scale models.
\begin{figure}
    \centering
    \includegraphics[width=0.5\linewidth]{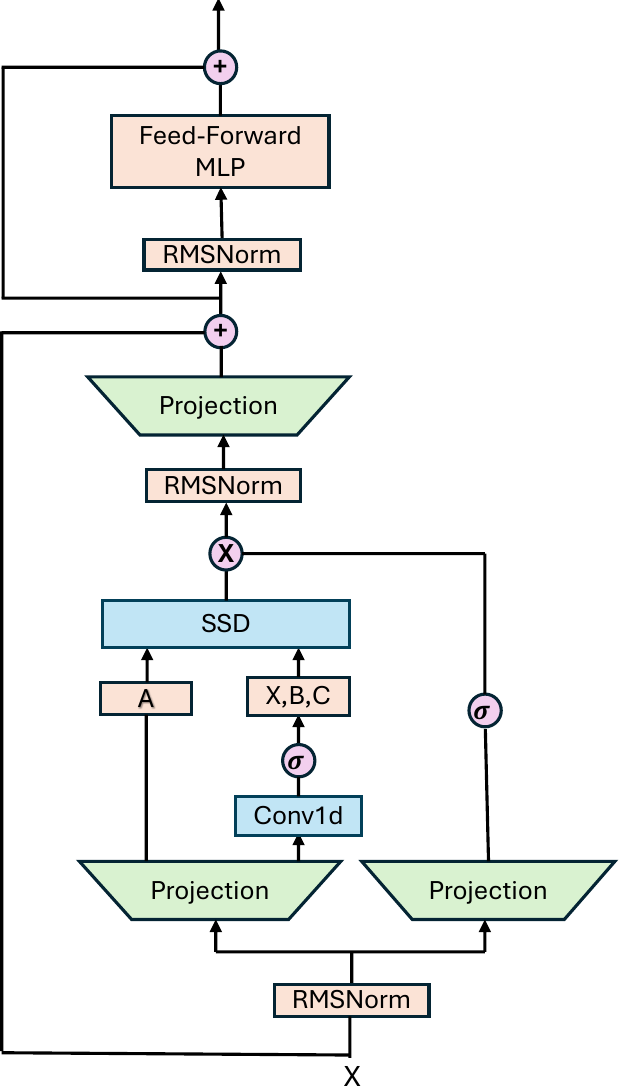}
    \caption{\JambaDNA{} Mamba2 Block}
    \label{fig:mamba2}
\end{figure}

\subsection{Pretraining dataset}
We utilized a comprehensive dataset comprising approximately 200 billion tokens, following the Nucleotide Transformer’s multi-species dataset \cite{dalla2023nucleotide}. It comprises of a subset of the NCBI dataset with 850 species and the details have been presented in the methodology section.

\subsection{Effectiveness of hybrid models}

\begin{figure}[t]
\centering
\includegraphics[width=0.9\textwidth]{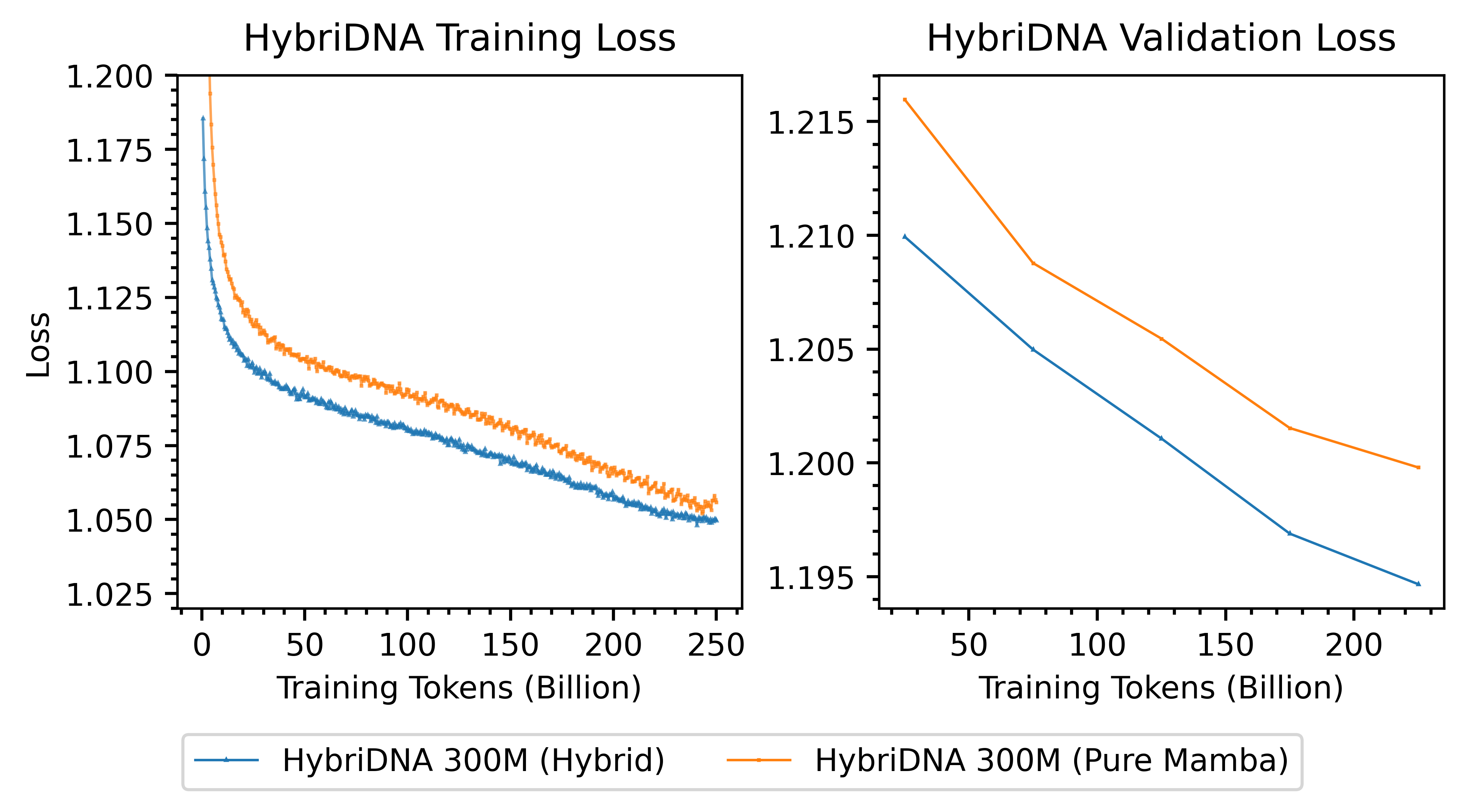}
\caption{Effectivenss of Hybridization in \JambaDNA{}}
\label{fig:Mamba_Jamba_Comp}
\end{figure}

To evaluate the effectiveness of incorporating Transformer layers alongside Mamba2 layers in the \JambaDNA{} model, we pretrained a 300M-size variant without any Transformer layers. 
Both our Hybrid model and the pure Mamba2 model were pretrained using an 8k context length.
Fig.~\ref{fig:Mamba_Jamba_Comp} presents the training and validation losses during the pretraining stage.
 It is evident that for models with similar parameter sizes, the hybrid model demonstrates lower training and validation losses compared to a model composed entirely of Mamba2 blocks.

\subsection{Fine-tuning configurations for downstream Tasks}\label{appendix:finetune_config}

In this section, we provide detailed fine-tuning procedures for each downstream dataset.

\textbf{GUE} For the GUE benchmark, we adhered to the specific settings for each task, including the warmup steps and training/validation steps that are customized for each task. 
The only adjustment we've made was to the learning rate: $5\mathrm{e}{-5}$ for the 300M model, $3\mathrm{e}{-5}$ for the 3B model, and $1\mathrm{e}{-5}$ for the 7B model. 
We applied a simple classification head for the model, using either the hidden state of the last token for classification in the standard setting, or the averaged hidden state of the repeated sequence input for the echo embedding setting.

\textbf{BEND} For the BEND dataset, we applied the exact settings across all three tasks, with a learning rate of $3\mathrm{e}{-3}$ and training for 100 epochs. 
The learning rate decreases linearly to 0, and we use the epoch with the lowest validation loss for the final test-set evaluation. 
In BEND, the model's embedding input is frozen, and only a downstream two-layer CNN model is fine-tuned for classification. 
The extracted hidden state is the average of hidden states of the input in the standard setting and the average of the repeated sequence in the echo-embedding setting. 

\textbf{Genomics LRB} Following the Genomics LRB work~\citep{trop2025the}, we fine-tune all the parameters of the model for all fine-tuning tasks, using a subsequent MLP for classification. We adhere to the benchmark's settings by inputting sequences with the pretraining context length into the model and averaging across the same length window for the same task across different models. For zero-shot tasks, we employ sequence-level probability for a regression correlation coefficient analysis, consistent with the benchmark's method.

\textbf{regLM} For the HyenaDNA baseline, we used the exact setting of the regLM model to load its fine-tuned checkpoint. 
For fine-tuning our \JambaDNA-300M model, we conducted 16 epochs on the human enhancer task with a learning rate of $1e-4$. 
For the Yeast Promoter task, as our model has already been pretrained on multi-species data, including yeast sequence, we fine-tuned it on the dataset for only 2 epochs, with the same learning rate of $1e-4$. 
Validation was performed every 400 steps for each task, and we saved the model with the highest validation accuracy as the final model. 
During generation, we use beam search with a beam size of 2500 for the human enhancer task and a beam size of 256 for the yeast promoter task to align with the total number of generated sequence of the original regLM setting.
The activity scoring models used were the same as those in the original regLM models.
The diversity metric is calculated as the mean pair-wise edit distance of the top-100 activity sequence across all labels for each task.

\section{GUE Benchmark Results}
\label{appendix:benchmark_gue}

% Table 1
\begin{table}[ht]
\centering
\resizebox{\textwidth}{!}{%
\begin{tabular}{llcccccc}
\toprule
\textbf{Model Type} & \textbf{Model} & \multicolumn{6}{c}{\textbf{Transcription Factor Prediction (Human)}} \\
\cmidrule(lr){3-8}
 & & \textbf{0} & \textbf{1} & \textbf{2} & \textbf{3} & \textbf{4} &  \\
 & &\emph{(MCC)} & \emph{(MCC)} & \emph{(MCC)} & \emph{(MCC)} & \emph{(MCC)}\\
\midrule
\multirow{5}{*}{\centering \textbf{Baselines}} & DNABERT-2 & 70.89 & 74.49 & 66.62 & 60.35 & 71.21 &  \\
 & NT-2.5B-MS & 66.46 & 70.25 & 58.70 & 51.28 &69.34 \\
 & NT-500M-human & 60.03 & 69.34 & 47.02 & 39.27 & 58.84 &  \\
 & Caduceus-Ph-131k & 70.69 & 69.00 & 61.13 & 55.98 & 69.07 &  \\
& HyenaDNA-160k & 64.47 & 70.74 & 60.44 & 39.78 & 73.27 &  \\
\midrule
\multirow{6}{*}{\textbf{Our Model}} & \JambaDNA-300M & 68.12 & 67.13 & 70.29 & 55.52 &80.80 \\
& \JambaDNA-300M(E) & 67.64 & 71.28 & 70.84 & 57.92 & 80.80 \\
& \JambaDNA-3B & 69.88 & 69.24 & 72.21 & 56.44 & 84.61 \\
& \JambaDNA-3B(E) & 69.02 & 70.82 & \textbf{72.80} & 58.01 & 85.02 \\
& \JambaDNA-7B & 70.00 & 74.47 & 70.42 & 64.52 & 85.03 \\
& \JambaDNA-7B(E) & \textbf{71.46} & \textbf{75.60} & 71.81 &\textbf{65.82} & \textbf{86.20} \\
\bottomrule
\end{tabular}%
}
\caption{Results for Transcription Factor Prediction (DNABERT2-Human) in the GUE benchmark}
\label{tab:tf_prediction}
\end{table}

% Table 2
\begin{table}[ht]
\centering
\resizebox{\textwidth}{!}{%
\begin{tabular}{llcccc}
\toprule
\textbf{Model Type} & \textbf{Model} & \multicolumn{3}{c}{\textbf{Promoter Detection (Human)}} & \textbf{Splice Site Prediction (Human)} \\
\cmidrule(lr){3-5} \cmidrule(lr){6-6}
 & & \textbf{all} & \textbf{notata} & \textbf{tata} & \textbf{reconstruct} \\
 & &\emph{(MCC)} & \emph{(MCC)} &\emph{(MCC)}&\emph{(MCC)}\\
\midrule
\multirow{5}{*}{\textbf{Baselines}} & DNABERT-2 &86.64 & 94.20 & 71.04 & 85.42 \\
 & NT-2.5B-MS & \textbf{91.00} & 94.02& \textbf{79.43} & 89.35 \\
 & NT-500M-human & 81.34 & 88.73 & 78.82 & 78.63 \\
 & Caduceus-Ph-131k & 83.98 & 92.13 & 70.96 & 71.80 \\
& HyenaDNA-160k & 83.04 & 91.03 & 66.36 & 77.76 \\
\midrule
\multirow{6}{*}{\textbf{Our Model}} & \JambaDNA-300M & 88.94& 94.44 & 69.63  & 87.74 \\
& \JambaDNA-300M(E) &88.81 &94.45&  68.45  & 88.72 \\
& \JambaDNA-3B & 89.48 & 94.49& 72.24 & 89.01 \\
& \JambaDNA-3B(E) & 89.30 & 94.33 & 73.02 & 89.10 \\
& \JambaDNA-7B & 88.28 & 94.73 & 73.59 & 90.09 \\
& \JambaDNA-7B(E) & 90.20 & \textbf{94.57} & 76.84 & \textbf{90.12}\\
\bottomrule
\end{tabular}%
}
\caption{Results for Promoter Detection and Splice Reconstruct (DNABERT2-Human) in the GUE benchmark}
\label{tab:model_comparison3}
\end{table}

% Table 3
\begin{table}[ht]
\centering
\resizebox{\textwidth}{!}{%
\begin{tabular}{llccccc}
\toprule
\textbf{Model Type} & \textbf{Model} & \multicolumn{3}{c}{\textbf{Core Promoter Detection (Human)}} \\
\cmidrule(lr){3-5} 
 & & \textbf{all} & \textbf{notata} & \textbf{tata}  \\
 & &\emph{(MCC)} & \emph{(MCC)} & \emph{(MCC)} \\
\midrule
\multirow{5}{*}{\textbf{Baselines}} & DNABERT-2 & 69.97 & 69.62 & \textbf{75.83} \\
 & NT-2.5B-MS & \textbf{70.28} & 71.49 & 72.95 \\
 & NT-500M-human & 63.36 & 64.67 & 72.34 \\
 & Caduceus-Ph-131k & 64.09 & 68.35 & 68.65 \\
& HyenaDNA-160k & 66.18 & 67.41 & 74.07  \\
\midrule
\multirow{6}{*}{\textbf{Our Model}} & \JambaDNA-300M &68.40 & 69.12&  69.09   \\
& \JambaDNA-300M(E)  & 68.37 & 69.15& 72.36  \\
& \JambaDNA-3B & 68.98 & 69.63& 69.89  \\
& \JambaDNA-3B(E)  & 68.90 & 70.01 & 73.21 \\
& \JambaDNA-7B & 66.50 & 70.66 & 76.94 \\
& \JambaDNA-7B(E) &  67.10&\textbf{71.53} &\textbf{77.49}\\
\bottomrule
\end{tabular}%
}
\caption{Results for Core Promoter Detection (DNABERT2-Human) in the GUE benchmark}
\label{tab:model_comparison_core_splice}
\end{table}

% Table 4
\begin{table}[ht]
\centering
\resizebox{\textwidth}{!}{%
\begin{tabular}{llccccccc}
\toprule
\textbf{Model Type} & \textbf{Model} & \multicolumn{6}{c}{\textbf{Transcription Factor Prediction (Mouse)}} &\textbf{Classification (Virus)} \\
\cmidrule(lr){3-8} \cmidrule(lr){9-9}
 & & \textbf{0} & \textbf{1} & \textbf{2} & \textbf{3} & \textbf{4} & &\textbf{Covid} \\
 & &\emph{(MCC)} & \emph{(MCC)} & \emph{(MCC)} & \emph{(MCC)} & \emph{(MCC)}&&\emph{(F-1)}\\
\midrule
\multirow{5}{*}{\textbf{Baselines}} & DNABERT-2 &56.76  & 84.77  &79.32  & 66.47 & 52.66 & & 71.02\\
 & NT-2.5B-MS & 63.31 & 83.76 & 71.52 & 69.44 & 47.07 & & 73.04 \\
 & NT-500M-human &31.04 &75.04 &61.67  &29.17  &29.27 & &50.82 \\
 & Caduceus-Ph-131k &50.44  & 82.63 &73.81  & 61.13 &43.40  & &40.35 \\
& HyenaDNA-160k &56.25  & 80.46 & 78.14 &60.83  & 46.25& &25.88 \\
\midrule
\multirow{6}{*}{\textbf{Our Model}} & \JambaDNA-300M &68.57 & 83.46  & 86.02 & 87.96 & 50.58 & &73.81  \\
& \JambaDNA-300M(E)  & 68.66 & 85.62 & 85.39 & 87.78 & 51.20 && 73.90  \\
& \JambaDNA-3B & 70.96 & 84.18 & 89.63 & \textbf{88.59} & 50.97 && 74.05  \\
& \JambaDNA-3B(E)  &  71.02 & 84.30 & \textbf{89.78} & 88.20 & 52.39 & &\textbf{74.88} \\
& \JambaDNA-7B & 71.68 & 87.75 & 86.59 & 87.62 & 56.47 && 74.02  \\
& \JambaDNA-7B(E) & \textbf{72.91} & \textbf{88.64} & 87.64 & \textbf{88.59} &\textbf{57.33} & &74.30 \\
\bottomrule
\end{tabular}%
}
\caption{Results for Transcription Factor Prediction (DNABERT2-Mouse) and Covid Variant Classification (DNABERT2-Virus) in the GUE benchmark}
\end{table}

% \section{Section title of first appendix}\label{secA1}

% An appendix contains supplementary information that is not an essential part of the text itself but which may be helpful in providing a more comprehensive understanding of the research problem or it is information that is too cumbersome to be included in the body of the paper.




\end{appendices}

\newpage
\bibliography{sn-article}% common bib file

\end{document}